\documentclass{article} 

\usepackage{iclr2026_conference,times}

\usepackage{amsmath,amssymb,amsthm,mathtools}
\usepackage{bm}
\usepackage{nicefrac}

\usepackage{graphicx}
\usepackage{caption}
\usepackage{subcaption}
\usepackage{wrapfig}

\usepackage{booktabs}
\usepackage{multirow}
\usepackage{makecell}
\usepackage{adjustbox}
\usepackage{float}

\usepackage[table]{xcolor}
\usepackage{colortbl}

\usepackage[ruled,lined]{algorithm2e}

\usepackage{microtype}
\usepackage{soul}
\usepackage{enumitem}
\usepackage{chngcntr}
\usepackage{placeins}

\usepackage[pagebackref=true,breaklinks=true,colorlinks,
            citecolor=acadblue,
            linkcolor=acadnavy,
            urlcolor=acadteal]{hyperref}
\usepackage{cleveref}
\usepackage{url}

\definecolor{acadnavy}{HTML}{1B3D6F}   
\definecolor{acadblue}{HTML}{2E6DA4}   
\definecolor{acadteal}{HTML}{1A7EB8}   
\definecolor{acadrowblue}{HTML}{EBF4FC} 
\definecolor{acadheadblue}{HTML}{D0E8F5}
\definecolor{acadbestgold}{HTML}{FFF0B3}
\definecolor{acadfocusgold}{HTML}{FFF8DE}
\definecolor{codepurple}{rgb}{0.58,0,0.82}
\definecolor{codegray}{rgb}{0.5,0.5,0.5}

\providecommand{\bestcell}[1]{\cellcolor{acadbestgold}\textbf{#1}}
\providecommand{\primaryhead}[1]{\cellcolor{acadfocusgold}\textbf{#1}}
\providecommand{\bluecell}[1]{\cellcolor{acadrowblue}#1}

\theoremstyle{plain}

\theoremstyle{definition}

\theoremstyle{remark}

\counterwithin{theorem}{section}
\counterwithin{proposition}{section}
\counterwithin{lemma}{section}
\counterwithin{corollary}{section}
\counterwithin{definition}{section}
\counterwithin{assumption}{section}
\counterwithin{remark}{section}

\iclrfinalcopy 
\title{A Good Initialization is All You Need for Faithful Visual Attribution}

\author{
\textbf{Zihan Gu}$^{1,2}$, \textbf{Jiayu Wang}$^{1,2}$, \textbf{Hua Zhang}$^{1,2,}$ , \textbf{Yue Hu}$^{1,2}$ \\
  $^{1}$Institute of Information Engineering, Chinese Academy of Sciences; \\ $^{2}$School of Cyber Security, University of Chinese Academy of Sciences; \\
  \texttt{\{guzihan,wangjiayu,zhanghua,huyue\}@iie.ac.cn}
}

\begin{document}

\maketitle

\begin{abstract}
Faithful visual attribution identifies which image regions support a model
prediction. Search-based perturbation methods lead the insertion--deletion
faithfulness frontier by masking regions and measuring score changes, but they
usually output a complete ordering of all regions. Many applications, especially
MLLM attribution and repair, only need a compact top-\(k\) evidence mask. We study
this mask-first attribution problem.
An exactly \(k\)-region mask is combinatorial: useful evidence can depend on
interactions among fine regions. Coarse grouping can stabilize early search but
aggregates redundant content, whereas one-step scoring can miss high-value
combinations. We introduce two forward-only methods. \textsc{CoPAIR} uses a
PhaseWin--Greedy gap diagnosis to construct coarse singleton/pair candidates that
warm-start full-ordering search. \textsc{TRACE} directly searches
fixed-cardinality fine-region masks with cross-entropy sampling, elite retention,
and distribution updates, with a finite-budget recovery analysis. The resulting
evidence set can be returned as a compact attribution mask or used to initialize
Greedy or PhaseWin when a complete ranking is required.
Across ImageNet classification with CLIP ViT-L/14, CLIP RN101, and ResNet-101,
our initialized search methods establish a new state-of-the-art frontier for
faithful full-ordering attribution under inclusive forward-call accounting. On
POPE and RePOPE with Qwen2.5-VL-3B-Instruct and LLaVA-v1.5-7B,
\textsc{TRACE}+Greedy gives the strongest search-based MLLM attribution results.
Direct \textsc{TRACE} masks further achieve single-point RePOPE repair rates of
\(94.44\%\) and \(96.00\%\), showing that compact evidence masks can be
actionable attribution outputs, not merely prefixes of full rankings.
\end{abstract}


\section{Introduction}
\begin{figure}[t!]
  \centering
  \includegraphics[width=\linewidth]{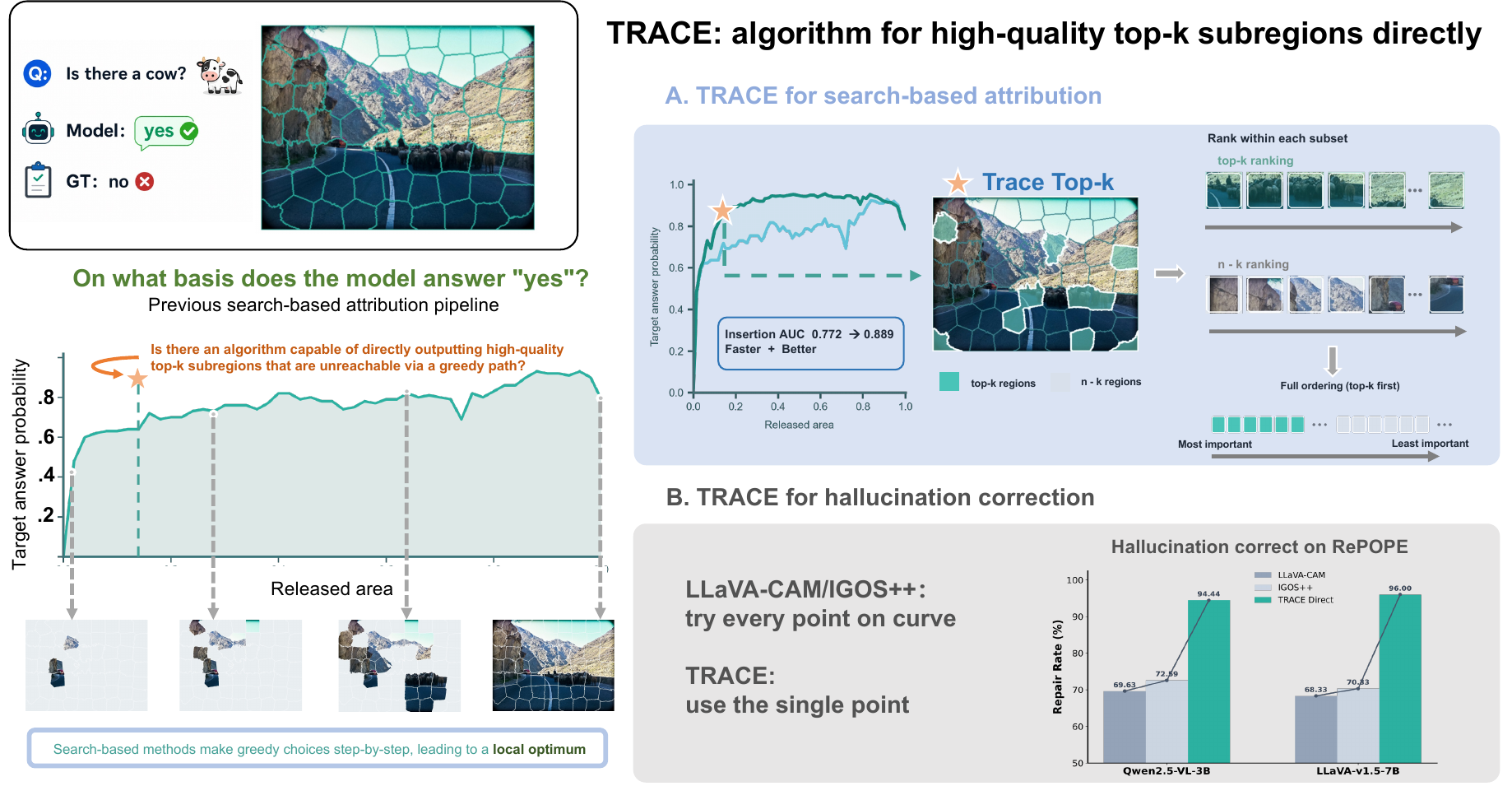}
  \caption{\textbf{From stepwise attribution curves to actionable top-\(k\) evidence.}
A conventional search-based pipeline constructs a complete insertion ordering for the
MLLM's hallucinated ``yes'' answer, but the greedy path may miss the best compact
evidence set. \textsc{TRACE} directly finds a top-\(k\) evidence anchor that either
initializes a top-\(k\)-first ordering, improving insertion AUC from 0.772 to 0.889, or
serves as a single repair intervention. Direct \textsc{TRACE} masks repair 94.44\% and
96.00\% of RePOPE cases for Qwen2.5-VL-3B-Instruct and LLaVA-v1.5-7B.}
  \label{fig:intro}
  \vspace{-24pt}
\end{figure}
Faithful visual attribution asks which image regions actually support a model
prediction.
Because explanations are used as evidence of model reliance, a plausible heatmap is
insufficient: faithfulness should be tested by intervening on the image and measuring
how the model response changes.
Insertion and deletion curves provide this operational test by asking whether selected
regions recover the prediction and whether removing them destroys it
~\citep{petsiuk2018rise,chen2024less,chen2025vps}.

Most strong attribution benchmarks assume a full-ordering contract: a method ranks all
image regions so that every prefix can be evaluated by insertion or deletion.
This is useful for full-curve evaluation, but many downstream uses only need a compact
evidence set.
In MLLM hallucination attribution and repair, the practical question is often not how
to rank every region, but which small set of visual evidence supports a generated
answer or intervention.
We therefore treat the fixed-size top-\(k\) evidence mask as a primary attribution
object, while retaining compatibility with full-ordering evaluation.

This mask-first contract changes the optimization problem.
The best \(k\)-region mask need not be a prefix of a greedy ranking, since its value can
depend on interactions among fine regions.
One-step scores can miss evidence that is only useful in combination, while coarse
grouping can stabilize search but may aggregate redundant content.
The challenge is to find a sparse, high-value fine-region evidence set under a
black-box, forward-only evaluation budget.

Search-based perturbation is the right foundation for this goal because it evaluates
masked inputs and directly optimizes quantities aligned with faithfulness, unlike
gradient-based explanations such as Grad-CAM~\citep{selvaraju2020grad} and Integrated
Gradients~\citep{sundararajan2017axiomatic}.
LIMA~\citep{chen2024less} showed that Greedy region search gives strong insertion
faithfulness, and later work extended this view to object-level foundation models
~\citep{chen2025vps} and autoregressive MLLM token generation~\citep{chen2026eagle}.
However, Greedy costs \(O(n^2)\) forward passes.
PhaseWin~\citep{gu2026phasewin} reduces this cost, but its extended analysis shows that
the residual gap to Greedy concentrates on partition-sensitive samples
~\citep{gu2026phasewinExtended}.
This suggests that early evidence quality is also a granularity problem: fine regions
can fragment evidence that would be more stable at a coarser scale.
This diagnosis motivates \textsc{CoPAIR} (Coarse Pairwise Attribution Initializer with
Release).
\textsc{CoPAIR} clusters fine regions into coarse spatial groups, evaluates strong
coarse singletons and pairs, and releases the selected candidate back into fine-region
continuation search.
It is therefore a principled initializer for Greedy or PhaseWin derived from the
PhaseWin--Greedy gap.
Its limitation is the same aggregation that makes it useful: a coarse group can merge
object parts, background fragments, and redundant fine regions, so it is not a minimal
compact explanation.
Thus \textsc{CoPAIR} improves full-ordering initialization, but does not solve the
mask-first problem.

We therefore introduce \textsc{TRACE} (Top-\(k\) Region Attribution via Cross-Entropy),
a forward-only method that directly optimizes fixed-cardinality fine-region masks.
\textsc{TRACE} samples \(k\)-hot candidate masks, evaluates them by the model response,
keeps elite samples, and updates the sampling distribution toward high-scoring evidence
combinations.
The returned mask can be used directly as the attribution object, or used to initialize
Greedy or PhaseWin when a complete ordering is required.
We also provide a finite-budget recovery analysis showing how the cross-entropy updates
increase probability mass on near-optimal evidence basins.
Figure~\ref{fig:intro} illustrates this shift from stepwise full-curve attribution to an
actionable top-\(k\) evidence anchor.

Empirically, \textsc{CoPAIR} and \textsc{TRACE} establish a new state-of-the-art frontier
for faithful visual attribution under a black-box, forward-only interface.
On ImageNet-derived classification splits with SLICO-64 regions
~\citep{achanta2012slic}, CLIP ViT-L/14, CLIP RN101~\citep{radford2021learning}, and
ResNet-101~\citep{he2016deep}, our initialized search methods improve task-relevant
full-ordering faithfulness over raw search under inclusive forward-call accounting.
On POPE~\citep{li2023evaluating} and RePOPE~\citep{neuhaus2025repope} with
Qwen2.5-VL-3B-Instruct~\citep{bai2025qwen25vl} and LLaVA-v1.5-7B
~\citep{liu2023improved}, \textsc{TRACE}+Greedy gives the strongest search-based MLLM
attribution results.
Direct \textsc{TRACE} masks further repair \(94.44\%\) and \(96.00\%\) of RePOPE cases,
respectively, using a single mask intervention rather than a full repair curve.

Our contributions are:
\begin{itemize}[leftmargin=1.5em,nosep]
  \item We formulate compact top-$k$ evidence masks as a primary output contract for
        faithful visual attribution, while preserving full-ordering evaluation.
  \item We derive \textsc{CoPAIR} from the PhaseWin--Greedy gap diagnosis as a
        coarse-pair initializer, and address its aggregation limitation with
        \textsc{TRACE}, a cross-entropy search method for fixed-size fine-region masks.
  \item We establish a new state-of-the-art frontier for forward-only faithful visual
        attribution across ImageNet classification, POPE, and RePOPE, and show that
        direct \textsc{TRACE} masks enable effective single-point MLLM repair.
\end{itemize}
\section{Related Work}
\label{sec:related}

\paragraph{Visual attribution methods.}
Gradient and activation saliency methods~\citep{simonyan2014deep,zeiler2014visualizing,
bach2015pixel,springenberg2015striving,smilkov2017smoothgrad,sundararajan2017axiomatic,
zhou2016learning,selvaraju2020grad,chattopadhyay2018gradcampp,wang2020score,
xie2023vitcx,zhao2024gradient,yamauchi2024spatial,zhao2024odam,li2025tam} are
computationally efficient but require white-box access and do not directly measure the
causal relationship between input regions and model outputs, a concern also reflected
in saliency sanity checks~\citep{adebayo2018sanity}.
Perturbation-based methods~\citep{ribeiro2016why,fong2017interpretable,
fong2019understanding,petsiuk2018rise,petsiuk2021black,novello2022making} and
Shapley-style attribution~\citep{shapley1953value,lundberg2017unified,kumar2020problems,
gudovskiy2018explain,sun2023explain} evaluate the model on masked inputs and are
therefore applicable to any black-box model, but achieve only moderate faithfulness due
to their unstructured or approximate nature.
Search-based attribution~\citep{chen2024less,chen2025vps} achieves the highest faithfulness
by casting region ranking as combinatorial optimization: a greedy algorithm iteratively selects
the region with the greatest marginal score gain, producing a provably high-quality ordering
at $O(n^2)$ cost. PhaseWin~\citep{gu2026phasewin,gu2026phasewinExtended} reduces this to $O(n)$ via
partition-dominant structure, but the remaining gap to Greedy concentrates on
partition-sensitive tail cases.
Our work targets a complementary object: a compact evidence mask that can be returned
directly when no full ranking is needed, or used as the initial state for Greedy and
PhaseWin when a complete ordering is required.

\paragraph{Attribution tasks.}
The dominant benchmark for visual attribution is image classification on
ImageNet~\citep{deng2009imagenet} with CNN, ViT, and CLIP backbones~\citep{he2016deep,
dosovitskiy2021image,radford2021learning}, where attribution identifies which image
regions drive the classifier's top-class prediction.
Faithful attribution for multimodal large language models (MLLMs) and adjacent
vision-language systems such as LLaVA-v1.5-7B~\citep{liu2023visual,liu2023improved},
BLIP-2~\citep{li2023blip2}, Grounding DINO~\citep{liu2024grounding},
Florence-2~\citep{xiao2024florence}, and
Qwen2.5-VL-3B-Instruct~\citep{bai2025qwen25vl} is far less explored, despite recent
MLLM explanation work~\citep{zhang2025mllms,xing2025where,li2025tam,
zhang2025redundancy,khorram2021igos++}.
Because answer generation depends on the interplay of visual tokens and language
context, gradient signals through the visual encoder are unreliable proxies for causal
influence. Defining the attribution score as the first-token probability of the generated
answer makes perturbation-based search directly applicable, but the $O(n^2)$ cost of
Greedy has historically made this impractical at scale.
EAGLE~\citep{chen2026eagle} adapts greedy search to autoregressive MLLM attribution and
introduces hallucination attribution and repair evaluation on POPE and
RePOPE~\citep{li2023evaluating,neuhaus2025repope}.
\textsc{TRACE} keeps the same black-box scoring interface while directly searching for a
fixed-size visual evidence mask; this mask can either support repair directly or warm
start the full search protocol.

\section{Method}
\label{sec:method}

\subsection{Search-Based Attribution Setup}
\label{subsec:search-setup}

We build on the search-based visual attribution framework of
Greedy~\citep{chen2024less}, VPS~\citep{chen2025vps}, and
PhaseWin~\citep{gu2026phasewin}. The input image is partitioned into fine
regions
\[
    U=\{u_1,\ldots,u_n\}.
\]
For a target \(y\), let \(f_y(x)\in[0,1]\) denote the normalized model response
on image \(x\), such as a class probability or an MLLM answer-token probability
under a fixed prompt. For any subset \(S\subseteq U\), its retained image is
\begin{equation}
    x_S = m(S)\odot x + \bigl(1-m(S)\bigr)\odot x_0 ,
    \label{eq:masked-image}
\end{equation}
where \(m(S)\) is the binary mask induced by \(S\) and \(x_0\) is a baseline
image. We use the same proxy family as prior faithful attribution search:
\begin{equation}
    F(S)=\operatorname{Suff}(S)
    \quad \text{or} \quad
    F(S)=\alpha \operatorname{Suff}(S)
    +(1-\alpha)\operatorname{Necc}(S),
    \label{eq:proxy}
\end{equation}
where
\[
    \operatorname{Suff}(S)=f_y(x_S),
    \qquad
    \operatorname{Necc}(S)=1-f_y(x_{U\setminus S}).
\]

A full-ordering attribution method returns a permutation
\(\Pi=(\pi_1,\ldots,\pi_n)\) of regions, so any prefix
\(\{\pi_1,\ldots,\pi_t\}\) can be evaluated by insertion or deletion curves.
However, many downstream uses of attribution do not require the whole ordering. They need
a compact top-\(k\) evidence mask \(S\) with \(|S|=k\), for example to highlight
the visual evidence behind an MLLM token or to perform a single repair
intervention. We therefore separate two output contracts:
\emph{initialization for full-ordering search}, and \emph{direct fixed-size
fine-region mask attribution}. Furthermore, the greedy nature of seeking a local optimum at each step—while yielding excellent ranking performance—often results in a loss of the ability to explore the global optimum. If we can directly output a high-value top-k prefix, the attribution performance of search-based algorithms will also be correspondingly enhanced.

\textbf{Search with initialization framework.} Given an initialized fine-region prefix
\[
    \Pi_0=(\pi_1,\ldots,\pi_r),
    \qquad
    S_{\mathrm{init}}=\{\pi_1,\ldots,\pi_r\},
\]
a continuation algorithm \(\mathcal A\) completes the remaining ordering as
\begin{equation}
    \Pi_{\mathrm{tail}}
    =
    \mathcal A\bigl(F; S_{\mathrm{init}}, U\setminus S_{\mathrm{init}}\bigr),
    \qquad
    \Pi=\Pi_0\oplus \Pi_{\mathrm{tail}} .
    \label{eq:initialized-search}
\end{equation}
Regions not included in \(S_{\mathrm{init}}\) are released back to the live pool
and can be selected later by the continuation search.

\paragraph{Internal ordering and release.}
For any candidate set \(C\subseteq U\), we order its fine regions by greedy
marginal rescoring inside \(C\):
\begin{equation}
    \pi_t
    =
    \arg\max_{u\in C\setminus P_{t-1}}
    F(P_{t-1}\cup\{u\}),
    \qquad
    P_{t-1}=\{\pi_1,\ldots,\pi_{t-1}\}.
    \label{eq:internal-greedy}
\end{equation}
Given a partial-release ratio \(\gamma\in(0,1]\), we keep the shortest prefix
satisfying
\begin{equation}
    r
    =
    \min\left\{
    t:
    F(P_t)\ge \gamma F(C)
    \right\},
    \qquad
    S_{\mathrm{init}}=P_r.
    \label{eq:release-rule}
\end{equation}
The remaining regions \(C\setminus S_{\mathrm{init}}\) are released to
\(U\setminus S_{\mathrm{init}}\) before downstream search. 

\subsection{CoPAIR: Coarse Pairwise Attribution Initializer with Release}
\label{subsec:copair}

We derive our initial design strategy by contrasting the PhaseWin algorithm with
the greedy approach. As detailed in \citep[Appendix]{gu2026phasewinExtended}, the
greedy method's apparent advantage in handling the long tail can be reinterpreted
as repeated early-prefix recovery: when an unfavorable fine partition fragments
visual evidence, global rescoring lets the model recover decisive fragments as
the context evolves. Coarser partitions aggregate these fragments into more
stable attribution units, mitigating this tail behavior directly. This
observation motivates \textsc{CoPAIR} (Algorithm~\ref{alg:copair}), a
deterministic coarse-spatial initializer that evaluates coarse singleton and
pair candidates first, then integrates the selected evidence into the
fine-region continuation search.

Let \(\mathcal G=\{G_1,\ldots,G_c\}\) denote a deterministic centroid-based
coarse partition of the fine regions,
\begin{equation}
    G_i\subseteq U,\quad
    G_i\cap G_j=\emptyset\ (i\ne j),
    \qquad
    \bigcup_{i=1}^c G_i=U,
    \label{eq:coarse-partition}
\end{equation}
with coarse count
\begin{equation}
    c =
    \operatorname{clip}
    \bigl(\operatorname{round}(\beta\sqrt{|U|}),c_{\min},c_{\max}\bigr).
    \label{eq:copair-count}
\end{equation}

CoPAIR's working assumption is that, once local evidence has been absorbed
within coarse groups, the early cross-group value is dominated by singleton and
pairwise terms. Formally, for \(I\subseteq[c]\),
\begin{equation}
    F\Bigl(\bigcup_{i\in I}G_i\Bigr)
    =
    \phi_\emptyset
    +\sum_{i\in I}\phi_i
    +\sum_{\substack{i<j\\i,j\in I}}\phi_{ij}
    +\sum_{\substack{J\subseteq I\\ |J|\ge 3}}\phi_J,
    \label{eq:coarse-interactions}
\end{equation}
where the higher-order remainder \(\sum_{|J|\ge 3}\phi_J\) is assumed small for
the early coarse candidates the initializer considers. Pair enumeration is
therefore not an independent scoring heuristic but an explicit search over
coarse-level interactions.

\begin{wrapfigure}{r}{0.48\textwidth}
\DontPrintSemicolon \SetAlgoNoEnd
\SetAlFnt{\small} \SetAlCapFnt{\small} \setlength{\algomargin}{0.8em}
\vspace{-14pt}
\begin{algorithm}[H]
\DontPrintSemicolon
\SetAlgoNoEnd
\caption{\textsc{CoPAIR}: Coarse Pairwise Attribution Initializer with Release}
\label{alg:copair}
\KwIn{Fine regions \(U\), proxy \(F\), coarse count \(c\) (Eq.~\eqref{eq:copair-count}), pool size \(q\),
pair margin \(\delta\), sufficiency threshold \(\tau\), release ratio \(\gamma\)}
Cluster \(U\) into coarse groups \(\mathcal G=\{G_1,\ldots,G_c\}\)\tcp*[r]{Eq.~\eqref{eq:coarse-partition}}
Evaluate \(F(G_i)\) for all \(G_i\in\mathcal G\)\;
\(\mathcal P\gets \operatorname{TopQ}_{G_i\in\mathcal G}F(G_i)\)\tcp*[r]{pool, Eq.~\eqref{eq:copair-pool}}
\((a,b)\gets\arg\max_{G_i,G_j\in\mathcal P,\,i<j}F(G_i\cup G_j)\)\tcp*[r]{Eq.~\eqref{eq:copair-pair}}
\(G_s\gets\arg\max_{G_i\in\mathcal G}F(G_i)\)\;
\If{\(F(G_a\cup G_b)-F(G_s)<\delta \ \land\ F(G_s)\ge\tau\)}{
    \(C_0\gets G_s\)\tcp*[r]{Eq.~\eqref{eq:copair-fallback}}
}
\Else{
    \(C_0\gets G_a\cup G_b\)\tcp*[r]{Eq.~\eqref{eq:copair-fallback}}
}
\If{\(F(C_0)<\tau\)}{
    Absorb one additional coarse group by Eq.~\eqref{eq:copair-third-block}\;
}
Order \(C_0\) by Eq.~\eqref{eq:internal-greedy} and release by Eq.~\eqref{eq:release-rule}\;
\Return \((\Pi_0,S_{\mathrm{init}})\)\;
\end{algorithm}
\vspace{-16pt}
\end{wrapfigure}

CoPAIR evaluates every coarse singleton and retains the top-\(q\) candidates as
a high-sufficiency pool,
\begin{equation}
    \mathcal P
    =
    \operatorname{TopQ}_{G_i\in\mathcal G} F(G_i),
    \label{eq:copair-pool}
\end{equation}
then exhaustively searches pairs within that pool,
\begin{equation}
    (a,b)
    =
    \arg\max_{G_i,G_j\in\mathcal P,\, i<j}
    F(G_i\cup G_j).
    \label{eq:copair-pair}
\end{equation}
Writing \(G_s=\arg\max_{G_i\in\mathcal G}F(G_i)\), \(C_{\mathrm{pair}}=G_a\cup
G_b\), and \(\Delta=F(C_{\mathrm{pair}})-F(G_s)\), the pair-gain margin
\(\delta\) and sufficiency threshold \(\tau\) select the initial coarse
candidate
\begin{equation}
    C_0 =
    \begin{cases}
        G_s,
        & \Delta < \delta \ \land\ F(G_s)\ge\tau,\\
        C_{\mathrm{pair}},
        & \text{otherwise}.
    \end{cases}
    \label{eq:copair-fallback}
\end{equation}

If \(F(C_0)<\tau\), CoPAIR absorbs one additional coarse group by greedy
sufficiency improvement,
\begin{equation}
\begin{aligned}
    G_h
    =
    \arg\max_{G\in\mathcal P:\,G\cap C_0=\emptyset}
    F(C_0\cup G),
    \\
    C_0 \leftarrow C_0\cup G_h .
\end{aligned}
\label{eq:copair-third-block}
\end{equation}
applied at most once regardless of whether \(C_0\) initially holds one group
(\(G_s\)) or two (\(C_{\mathrm{pair}}\)), so the candidate never exceeds three
coarse groups.

Finally, CoPAIR applies the internal greedy ordering and release rules defined
in Eqs.~\eqref{eq:internal-greedy}--\eqref{eq:release-rule} to \(C_0\), yielding
\((\Pi_0, S_{\mathrm{init}})\) as a starting point for continuation search — the
return step of Algorithm~\ref{alg:copair}. CoPAIR thus serves as a structured
initializer for both Greedy and PhaseWin. Notably, it is not intended to solve
the fixed-\(k\) fine-region mask problem directly; that objective requires
stepping outside this specific search framework, a task the core TRACE
algorithm described below addresses directly.

\subsection{TRACE: Top-k Region Attribution via Cross-Entropy}
\label{subsec:trace}

\textsc{TRACE} (Algorithm~\ref{alg:trace}) directly targets the fixed-cardinality
fine-region mask problem. A candidate mask is represented by
\begin{equation}
\begin{aligned}
    z\in\{0,1\}^{|U|},
    \;
    \|z\|_0=k,
    \;
    S(z)=\{u_i:z_i=1\}.
    \label{eq:trace-mask}
\end{aligned}
\end{equation}

Unlike CoPAIR, which constructs a coarse candidate for continuation search,
TRACE searches the exactly-\(k\) fine-region mask space directly. The returned
mask is the best candidate observed under the finite forward budget, not a
certified global optimum; Appendix~\ref{app:trace-details} gives the
finite-budget recovery analysis and the notation used for this probabilistic
view.

TRACE adapts the cross-entropy method~\citep{rubinstein1999cross} to the
discrete \(k\)-hot space: it maintains logits \(\theta\in\mathbb R^{|U|}\),
samples candidate masks, retains elite candidates, and updates the
distribution toward regions that repeatedly appear in high-scoring masks.

\begin{wrapfigure}{r}{0.48\textwidth}
\DontPrintSemicolon \SetAlgoNoEnd
\SetAlFnt{\small} \SetAlCapFnt{\small} \setlength{\algomargin}{0.8em}
\vspace{-4pt}
\begin{algorithm}[H]
\DontPrintSemicolon
\SetAlgoNoEnd
\caption{\textsc{TRACE}: Top-\(k\) Region Attribution via Cross-Entropy}
\label{alg:trace}
\KwIn{Regions \(U\), proxy \(F\), mask size \(k\), rounds \(R\), samples \(M\),
elite ratio \(\rho_{\mathrm e}\), temperature \(T\), update rate \(\eta\),
smoothing \(\lambda_{\mathrm{sm}}\), clip \(\epsilon\), logit bound \(\theta_{\max}\)}
Initialize \(\theta_i=0\) for all \(u_i\in U\)\;
\(S^\star\gets\emptyset,\ b^\star\gets-\infty\)\;
\For{\(r=1,\ldots,R\)}{
    \For{\(m=1,\ldots,M\)}{
        Sample \(z^{(m)}\) by Eq.~\eqref{eq:trace-sampling}\;
        \(b^{(m)}\gets F(S(z^{(m)}))\)\tcp*[r]{Eq.~\eqref{eq:trace-score}}
        \If{\(b^{(m)}>b^\star\)}{
            \(S^\star\gets S(z^{(m)}),\ b^\star\gets b^{(m)}\)\;
        }
    }
    \(\mathcal E\gets\) top \(\lceil\rho_{\mathrm e}M\rceil\) samples by \(b^{(m)}\)\;
    Estimate \(\hat p_i\) by \(\hat p_i =
    \frac{1}{|\mathcal E|}
    \sum_{z\in\mathcal E} z_i .\)\;
    Smooth \(\hat p_i\to\tilde p_i\) by Eq.~\eqref{eq:trace-smooth}\;
    Update \(\theta_i\) by Eq.~\eqref{eq:trace-update}\;
}
Order \(S^\star\) by Eq.~\eqref{eq:internal-greedy} and release by Eq.~\eqref{eq:release-rule}\;
\Return direct mask \(S^\star\), and initializer \((\Pi_0,S_{\mathrm{init}})\)\;
\end{algorithm}
\vspace{20pt}
\end{wrapfigure}

At round \(r\), each candidate is sampled by Gumbel-top-\(k\)
perturbation~\citep{kool2019stochastic}: for sample \(m\), draw
\(g_i^{(m)}\sim\operatorname{Gumbel}(0,1)\) i.i.d.\ and select
\begin{equation}
\resizebox{0.8\linewidth}{!}{$
\begin{aligned}
    z_i^{(m)} =
    \mathbf{1}\left[
    i\in
    \operatorname{TopK}
    \left\{
    \theta_j/T+g_j^{(m)}
    \right\}_{j=1}^{|U|}
    \right],
    \label{eq:trace-sampling}
\end{aligned}
$}
\end{equation}
where \(T\) is the temperature. The candidate score is
\begin{equation}
    b^{(m)} = F(S(z^{(m)})).
    \label{eq:trace-score}
\end{equation}
Let \(\mathcal E\) be the top \(\lceil\rho_{\mathrm e}M\rceil\) samples by
\(b^{(m)}\), where \(\rho_{\mathrm e}\) is the elite ratio. The empirical elite
inclusion frequency is
\(
    \hat p_i =
    \frac{1}{|\mathcal E|}
    \sum_{z\in\mathcal E} z_i .
\)

Exact maximum-likelihood fitting of a Gumbel-top-\(k\) distribution to elite
sets has no closed-form componentwise update, so TRACE instead uses an
inclusion-moment surrogate: the smoothed marginal target
\begin{equation}
    \tilde p_i =
    (1-\lambda_{\mathrm{sm}})\hat p_i
    +\lambda_{\mathrm{sm}}\frac{k}{|U|},
    \label{eq:trace-smooth}
\end{equation}
drives the logit update
\begin{equation}
    \theta_i
    \leftarrow
    \operatorname{clip}
    \left(
    (1-\eta)\theta_i
    +
    \eta\operatorname{logit}
    \bigl(\operatorname{clip}(\tilde p_i,\epsilon,1-\epsilon)\bigr),
    -\theta_{\max},\theta_{\max}
    \right),
    \label{eq:trace-update}
\end{equation}
where \(\eta\) is the update rate, \(\epsilon\) prevents saturated logits, and
\(\theta_{\max}\) bounds the logits.

As shown in Algorithm~\ref{alg:trace}, TRACE tracks the best observed mask
incrementally as each candidate is scored, equivalently
\(
    S^\star
    =
    \arg\max_{S(z)\ \text{evaluated}} F(S(z)).
\)
When the task only needs a compact evidence set, \(S^\star\) is returned
directly as the final attribution mask. When a complete ordering is required,
TRACE applies the same internal greedy ordering and release rule from
Eqs.~\eqref{eq:internal-greedy}--\eqref{eq:release-rule} to \(S^\star\),
obtaining \((\Pi_0,S_{\mathrm{init}})\), and then uses the continuation
interface in Eq.~\eqref{eq:initialized-search}.

The nominal TRACE search evaluates \(RM\) candidate masks before duplicate
caching. Under the sufficiency-only proxy used in the main experiments, each
uncached candidate requires one model forward, and masks from the same round
can be evaluated as a batch. If the combined sufficiency--necessity proxy in
Eq.~\eqref{eq:proxy} is used instead, each uncached candidate requires both the
retained and complement evaluations. The default TRACE budget is thus
independent of the number of fine regions once \(R\), \(M\), and \(k\) are
fixed, while the continuation cost depends on the remaining live pool.
\section{Experiments}
\label{sec:experiments}
\providecommand{\metricsep}{\hspace{3pt}{\color{black!20}\vrule width 0.35pt}\hspace{3pt}}
We evaluate two attribution contracts. When a complete ordering is required,
\textsc{TRACE} and \textsc{CoPAIR} initialize Greedy (G) or PhaseWin (PW) and
are evaluated by insertion--deletion replay. When only a compact evidence set is
required, the selected mask is applied directly as a single repair intervention.
All tables report task-matched metrics and, for forward-only methods, inclusive
selection cost. All initializer hyperparameters, including CoPAIR coarse-pair settings, TRACE sampling settings, duplicate caching, internal greedy ordering, and the \(0.8\) partial-release rule, are reported in Appendix~\ref{app:init-hparams}.
Additional protocol details and table conventions are given in
Appendix~\ref{app:experimental-details}.
\subsection{Setup and Baselines}
\label{subsec:exp-setup}
For image classification, we follow the faithful visual attribution protocol used by
LIMA, VPS, and PhaseWin~\citep{chen2024less,chen2025vps,gu2026phasewin}. We evaluate
CLIP ViT-L/14 and CLIP RN101~\citep{radford2021learning}, and supervised
ResNet-101~\citep{he2016deep}, on ImageNet-derived Correct, Cause, and Repair
splits~\citep{deng2009imagenet} with SLICO-64 regions~\citep{achanta2012slic}.
The non-search classification baselines include RISE~\citep{petsiuk2018rise},
HSIC~\citep{novello2022making}, vanilla gradient saliency~\citep{simonyan2014deep},
Grad-ECLIP~\citep{zhao2024gradient}, IG2~\citep{sundararajan2017axiomatic}, and
IGOS++~\citep{khorram2021igos++}. For MLLMs, we evaluate
the POPE decision-faithfulness and RePOPE repair settings~\citep{li2023evaluating,neuhaus2025repope} with
Qwen2.5-VL-3B-Instruct~\citep{bai2025qwen25vl} and
LLaVA-v1.5-7B~\citep{liu2023improved}. The MLLM attribution and repair protocol
follows EAGLE~\citep{chen2026eagle}; its non-search references include
LLaVA-CAM~\citep{zhang2025redundancy} and IGOS++~\citep{khorram2021igos++}.

\subsection{Metrics and Accounting}
\label{subsec:metrics}
All scores use normalized target responses \(f_y(x)\in[0,1]\). For ordered
attribution, we report insertion AUC (Ins; higher is better), deletion AUC
(Del; lower is better), the insertion score at 50\% released area (@50), and
the maximum insertion score (High). Correct, Cause, and POPE emphasize Ins;
classification Repair and RePOPE emphasize High.
For single-point RePOPE repair, direct masks are evaluated by Repair Rate,
with Correct/Total and Failed reporting the corresponding counts. Full ordered
repair metrics such as AMCR, CSR@10, and CorrectionRate are deferred to
Appendix~\ref{app:experimental-details}.
MEC(k) reports thousands of selection-time model forwards and excludes
insertion/deletion replay. For initialized rows, MEC is inclusive: it includes
uncached initializer calls and downstream continuation calls. Under the default
sufficiency-only TRACE scoring, each uncached sampled mask uses one forward;
using the combined sufficiency--necessity proxy would require both retained and
complement evaluations. MEC is compared only within the forward-only family and
is left blank for gradient, CAM, and forward-backward optimization methods.

\subsection{Image Classification Attribution}
\label{subsec:classification-results}
\begin{table}[t!]
\centering
\vspace{-16pt}
\caption{\textbf{CLIP ViT-L/14 classification attribution on ImageNet-derived splits.} Correct and Cause emphasize \textbf{Ins}; Repair emphasizes \textbf{High}. Gold cells mark emphasized winners. G denotes Greedy and PW denotes PhaseWin.}
\label{tab:clip-vit-main}
\scriptsize
\setlength{\tabcolsep}{2.1pt}
\renewcommand{\arraystretch}{0.92}
\resizebox{\linewidth}{!}{
\begin{tabular}{@{}l@{\metricsep}*{5}{c}@{\metricsep}*{5}{c}@{\metricsep}*{5}{c}@{}}
\toprule
Method
& \multicolumn{5}{c}{Correct}
& \multicolumn{5}{c}{Cause}
& \multicolumn{5}{c}{Repair} \\
\cmidrule(lr){2-6}\cmidrule(lr){7-11}\cmidrule(l){12-16}
& \primaryhead{Ins} & Del & @50 & High & MEC(k)
& \primaryhead{Ins} & Del & @50 & High & MEC(k)
& Ins & Del & @50 & \primaryhead{High} & MEC(k) \\
\midrule
RISE & 0.6173 & 0.3103 & 0.8110 & 0.9325 & 2.00 & 0.4019 & 0.1484 & 0.5769 & 0.7551 & 2.00 & 0.1986 & 0.0458 & 0.3786 & 0.4555 & 2.00 \\
HSIC & 0.6624 & 0.2617 & 0.8398 & 0.9223 & 0.50 & 0.3806 & 0.1481 & 0.5752 & 0.7078 & 0.50 & 0.1369 & 0.0593 & 0.3286 & 0.3614 & 0.50 \\
Gradient & 0.4353 & 0.4750 & 0.5148 & 0.9118 & -- & 0.2555 & 0.2740 & 0.3253 & 0.6889 & -- & 0.0964 & 0.0924 & 0.1874 & 0.3172 & -- \\
Grad-ECLIP & 0.6420 & 0.2802 & 0.8067 & 0.9309 & -- & 0.3811 & 0.1603 & 0.5560 & 0.7264 & -- & 0.1818 & 0.0492 & 0.4117 & 0.4609 & -- \\
IG2 & 0.4083 & 0.4971 & 0.4610 & 0.9119 & -- & 0.2415 & 0.2836 & 0.2956 & 0.6892 & -- & 0.0915 & 0.1007 & 0.1739 & 0.3157 & -- \\
IGOS++ & 0.5018 & 0.3992 & 0.6265 & 0.9136 & -- & 0.2740 & 0.2247 & 0.3678 & 0.6878 & -- & 0.1114 & 0.0836 & 0.2229 & 0.3291 & -- \\
\midrule
Greedy (G) & 0.8431 & 0.1173 & 0.9597 & 0.9787 & 4.08 & 0.7237 & 0.0500 & 0.8853 & 0.9131 & 4.09 & 0.5448 & 0.0197 & 0.7749 & 0.8009 & 4.09 \\
PhaseWin (PW) & 0.8037 & 0.1510 & 0.9308 & 0.9707 & 1.73 & 0.6604 & 0.0563 & 0.8325 & 0.8953 & 2.00 & 0.4482 & 0.0206 & 0.6696 & 0.7298 & 2.00 \\
\rowcolor{acadrowblue}
\textsc{CoPAIR}+PW & \(0.8419^{+.0382}\) & 0.1617 & 0.9588 & 0.9736 & \(1.47_{-.26}\) & \(0.7034^{+.0430}\) & 0.0745 & 0.8717 & 0.9013 & \(1.79_{-.21}\) & 0.5152 & 0.0248 & 0.7688 & \(0.7867^{+.0569}\) & \(2.04_{+.03}\) \\
\rowcolor{acadrowblue}
\textsc{TRACE}+PW & \(0.8451^{+.0414}\) & 0.1631 & 0.9551 & 0.9736 & \(1.52_{-.20}\) & \(0.7193^{+.0589}\) & 0.0747 & 0.8821 & 0.9076 & \(1.83_{-.17}\) & 0.5241 & 0.0265 & 0.7757 & \(0.7961^{+.0663}\) & \(2.06_{+.06}\) \\
\rowcolor{acadrowblue}
\textsc{CoPAIR}+G & \(0.8625^{+.0194}\) & 0.1344 & 0.9692 & 0.9796 & \(3.60_{-.48}\) & \(0.7370^{+.0133}\) & 0.0680 & 0.8925 & 0.9132 & \(3.52_{-.57}\) & 0.5545 & 0.0226 & 0.7947 & \(0.8105^{+.0096}\) & \(3.60_{-.49}\) \\
\rowcolor{acadrowblue}
\textsc{TRACE}+G & \bestcell{\(\mathbf{0.8679}^{+.0248}\)} & 0.1342 & 0.9677 & 0.9799 & \(3.69_{-.39}\) & \bestcell{\(\mathbf{0.7527}^{+.0290}\)} & 0.0676 & 0.9003 & 0.9226 & \(3.63_{-.46}\) & 0.5632 & 0.0236 & 0.8007 & \bestcell{\(\mathbf{0.8190}^{+.0181}\)} & \(3.65_{-.44}\) \\
\bottomrule
\end{tabular}}
\vspace{-10pt}
\end{table}
\begin{figure}[t]
\centering
\includegraphics[width=0.8\linewidth]{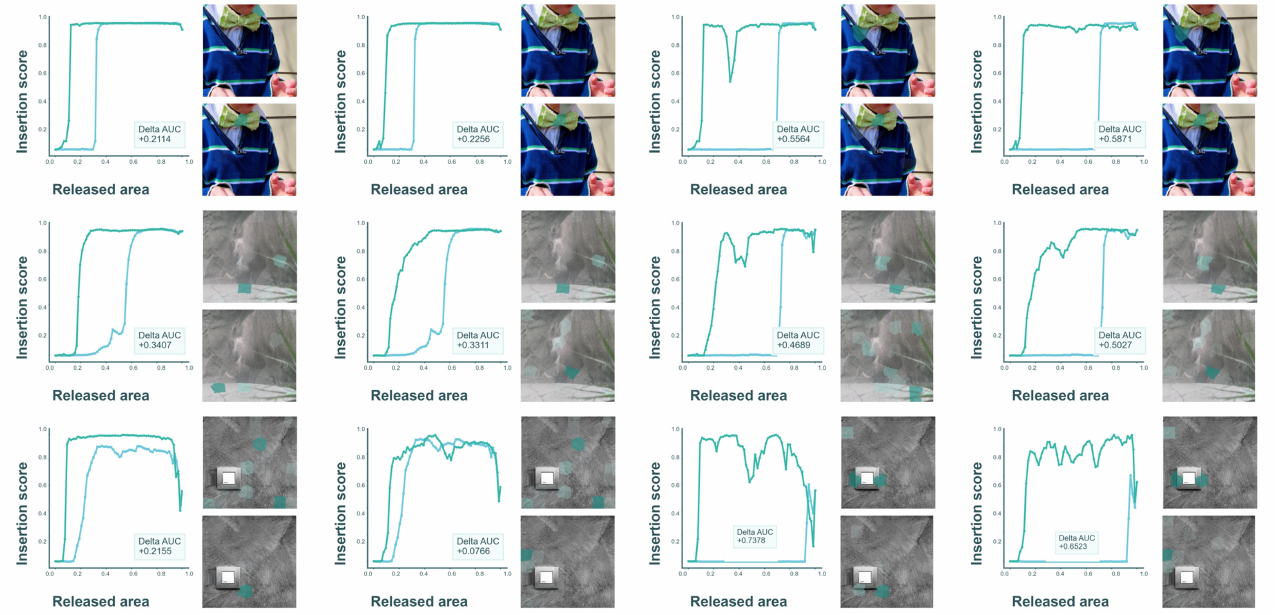}
\caption{\textbf{Qualitative CLIP ViT-L/14 Correct-split examples.} Each panel compares the insertion trajectory and the
visual evidence masks for a representative image; positive \(\Delta\)AUC indicates that
the initialized ordering recovers the target score earlier than the raw ordering.}
\label{fig:clip-vit-correct-vis}
\vspace{-12pt}
\end{figure}

Table~\ref{tab:clip-vit-main} gives the full CLIP ViT-L/14 landscape, including
perturbation, gradient, optimization-based, and search-based methods. Superscripts on initialized rows report the absolute gain over the corresponding raw search in the emphasized metric; MEC subscripts report the cost change against the same raw search.  The search
family is already much stronger than non-search baselines, but initialization still
moves the frontier. \textsc{TRACE}+G gives the best task-relevant score in all three
splits: Correct Ins improves from 0.8431 to 0.8679, Cause Ins from 0.7237 to 0.7527,
and Repair High from 0.8009 to 0.8190. These gains are not obtained by spending more
search: the inclusive MEC for \textsc{TRACE}+G is lower than raw Greedy in each split.
PW-based continuation also benefits from initialization, although its inclusive MEC
can increase slightly on Repair when the seed cost is not fully offset by the smaller
continuation problem.
Figure~\ref{fig:clip-vit-correct-vis} shows representative Correct-split cases behind
Table~\ref{tab:clip-vit-main}: the initialized curve recovers the class score earlier,
and the displayed masks concentrate the early evidence on the object regions that
support the prediction.

Table~\ref{tab:classification-search-main} checks whether the pattern depends on the
CLIP ViT-L/14 backbone. We only show the search family here, and defer the full
baseline landscape for CLIP RN101 and ResNet-101 to Appendix~\ref{app:classification-full}.
The result is useful precisely because it is not a single-method sweep. On CLIP RN101,
\textsc{CoPAIR}+G is the strongest variant for the emphasized metrics, suggesting that
coarse spatial pairing can be valuable when the backbone favors coarser evidence
units. On ResNet-101, \textsc{TRACE}+G is strongest on Correct Ins, Cause Ins, and
Repair High. Across both backbones, at least one initialized search variant improves
the emphasized metric over its raw continuation baseline, showing that initialization
changes the attribution search frontier rather than merely reformatting the same
ranking.

Furthermore, we conducted ablation studies on $k$ and various hyperparameters of TRACE using classification scenarios. We also performed comparative experiments to demonstrate the necessity of the probabilistic mask. These experimental results validate the robustness of TRACE. Detailed experimental results are provided in Appendix~\ref{app:additional-experiments}.

\begin{table}[t]
\centering
\vspace{-14pt}
\caption{\textbf{Search-based classification attribution on CLIP RN101 and ResNet-101 using ImageNet-derived splits.}}
\label{tab:classification-search-main}
\scriptsize
\setlength{\tabcolsep}{1.8pt}
\renewcommand{\arraystretch}{0.92}
\resizebox{\linewidth}{!}{
\begin{tabular}{@{}l@{\metricsep}*{5}{c}@{\metricsep}*{5}{c}@{\metricsep}*{5}{c}@{}}
\toprule
Method
& \multicolumn{5}{c}{Correct}
& \multicolumn{5}{c}{Cause}
& \multicolumn{5}{c}{Repair} \\
\cmidrule(lr){2-6}\cmidrule(lr){7-11}\cmidrule(l){12-16}
& \primaryhead{Ins} & Del & @50 & High & MEC(k)
& \primaryhead{Ins} & Del & @50 & High & MEC(k)
& Ins & Del & @50 & \primaryhead{High} & MEC(k) \\
\midrule
\rowcolor{black!8}
\multicolumn{16}{@{}c@{}}{\textit{CLIP RN101}} \\
Greedy (G) & 0.6677 & 0.0462 & 0.8243 & 0.9134 & 4.08 & 0.5163 & 0.0223 & 0.6655 & 0.7780 & 4.08 & 0.3195 & 0.0092 & 0.4739 & 0.5478 & 4.08 \\
PhaseWin (PW) & 0.5691 & 0.0500 & 0.6725 & 0.8886 & 1.79 & 0.4087 & 0.0230 & 0.4947 & 0.7366 & 1.83 & 0.2265 & 0.0090 & 0.3012 & 0.4657 & 1.52 \\
\rowcolor{acadrowblue}
\textsc{CoPAIR}+PW & \(0.6688^{+.0997}\) & 0.0783 & 0.8128 & 0.8975 & \(1.67_{-.12}\) & \(0.5019^{+.0932}\) & 0.0378 & 0.6483 & 0.7505 & \(1.90_{+.08}\) & 0.2983 & 0.0122 & 0.4633 & \(0.5170^{+.0513}\) & \(1.87_{+.35}\) \\
\rowcolor{acadrowblue}
\textsc{TRACE}+PW & \(0.6376^{+.0685}\) & 0.0750 & 0.7618 & 0.8928 & \(1.76_{-.03}\) & \(0.4753^{+.0666}\) & 0.0377 & 0.6000 & 0.7437 & \(1.92_{+.09}\) & 0.2908 & 0.0117 & 0.4246 & \(0.5073^{+.0416}\) & \(1.82_{+.30}\) \\
\rowcolor{acadrowblue}
\textsc{CoPAIR}+G & \bestcell{\(\mathbf{0.7038}^{+.0361}\)} & 0.0746 & 0.8535 & 0.9134 & \(3.39_{-.68}\) & \bestcell{\(\mathbf{0.5454}^{+.0291}\)} & 0.0360 & 0.7031 & 0.7824 & \(3.39_{-.70}\) & 0.3348 & 0.0119 & 0.5084 & \bestcell{\(\mathbf{0.5595}^{+.0117}\)} & \(3.43_{-.65}\) \\
\rowcolor{acadrowblue}
\textsc{TRACE}+G & \(0.6852^{+.0175}\) & 0.0718 & 0.8262 & 0.9102 & \(3.58_{-.50}\) & \(0.5260^{+.0097}\) & 0.0373 & 0.6725 & 0.7708 & \(3.56_{-.52}\) & 0.3344 & 0.0116 & 0.4913 & \(0.5563^{+.0085}\) & \(3.64_{-.44}\) \\
\midrule
\rowcolor{black!8}
\multicolumn{16}{@{}c@{}}{\textit{ResNet-101}} \\
Greedy (G) & 0.8407 & 0.1301 & 0.9554 & 0.9605 & 4.08 & 0.7390 & 0.0479 & 0.8958 & 0.9155 & 4.09 & 0.5482 & 0.0173 & 0.7604 & 0.7758 & 4.09 \\
PhaseWin (PW) & 0.8043 & 0.1483 & 0.9344 & 0.9462 & 1.75 & 0.6802 & 0.0536 & 0.8450 & 0.8883 & 1.87 & 0.4670 & 0.0185 & 0.6705 & 0.7067 & 1.80 \\
\rowcolor{acadrowblue}
\textsc{CoPAIR}+PW & \(0.8145^{+.0102}\) & 0.1630 & 0.9402 & 0.9469 & \(1.49_{-.26}\) & \(0.6971^{+.0169}\) & 0.0659 & 0.8665 & 0.8901 & \(1.67_{-.20}\) & 0.5017 & 0.0214 & 0.7341 & \(0.7479^{+.0412}\) & \(1.87_{+.07}\) \\
\rowcolor{acadrowblue}
\textsc{TRACE}+PW & \(0.8214^{+.0171}\) & 0.1581 & 0.9414 & 0.9487 & \(1.52_{-.22}\) & \(0.7035^{+.0233}\) & 0.0648 & 0.8603 & 0.8916 & \(1.69_{-.18}\) & 0.5120 & 0.0214 & 0.7358 & \(0.7511^{+.0444}\) & \(1.86_{+.06}\) \\
\rowcolor{acadrowblue}
\textsc{CoPAIR}+G & \(0.8394^{-.0013}\) & 0.1451 & 0.9551 & 0.9602 & \(3.65_{-.43}\) & \(0.7346^{-.0044}\) & 0.0606 & 0.8971 & 0.9142 & \(3.57_{-.51}\) & 0.5377 & 0.0206 & 0.7662 & \(0.7778^{+.0020}\) & \(3.63_{-.45}\) \\
\rowcolor{acadrowblue}
\textsc{TRACE}+G & \bestcell{\(\mathbf{0.8478}^{+.0071}\)} & 0.1420 & 0.9559 & 0.9616 & \(3.73_{-.35}\) & \bestcell{\(\mathbf{0.7414}^{+.0024}\)} & 0.0597 & 0.8892 & 0.9139 & \(3.64_{-.44}\) & 0.5490 & 0.0201 & 0.7671 & \bestcell{\(\mathbf{0.7794}^{+.0036}\)} & \(3.65_{-.44}\) \\
\bottomrule
\end{tabular}}
\vspace{-12pt}
\end{table}

\subsection{MLLM Hallucination Attribution}
\label{subsec:mllm-attribution-results}

Table~\ref{tab:mllm-main} keeps the full MLLM attribution landscape in the main text,
because hallucination attribution is the setting where compact evidence masks are
most directly actionable. The table groups results by model and dataset. LLaVA-CAM
and IGOS++ provide useful non-search references, but they are far below search-based
methods on insertion faithfulness. On POPE, \textsc{TRACE}+G gives the best Ins for
both Qwen2.5-VL-3B and LLaVA-v1.5-7B. On RePOPE, raw Greedy remains competitive in
Ins for Qwen, but the repair-relevant High metric is highest for \textsc{TRACE}+G on
both models. This distinction matters: RePOPE asks whether the selected evidence can
recover the corrected answer, so the peak confidence along the insertion trajectory is
the more direct signal for repair.

\begin{table}[ht]
\centering
\vspace{-8pt}
\caption{\textbf{MLLM attribution on POPE and RePOPE.} POPE evaluates decision faithfulness and emphasizes \textbf{Ins}; RePOPE evaluates repair-oriented evidence recovery and emphasizes \textbf{High}.}
\label{tab:mllm-main}
\scriptsize
\setlength{\tabcolsep}{3.2pt}
\renewcommand{\arraystretch}{0.85}
\resizebox{0.75\linewidth}{!}{
\begin{tabular}{@{}llcccccccc@{}}
\toprule
\multirow{2}{*}{Dataset} & \multirow{2}{*}{Method}
& \multicolumn{4}{c}{Qwen2.5-VL-3B}
& \multicolumn{4}{c}{LLaVA-v1.5-7B} \\
\cmidrule(lr){3-6}\cmidrule(l){7-10}
& & \primaryhead{Ins} & Del & \primaryhead{High} & MEC(k) & \primaryhead{Ins} & Del & \primaryhead{High} & MEC(k) \\
\midrule
\multirow{8}{*}{POPE}
& LLaVA-CAM & 0.7652 & 0.7736 & 0.9506 & -- & 0.7563 & 0.7735 & 0.9270 & -- \\
& IGOS++ & 0.7754 & 0.7542 & 0.9504 & -- & 0.7879 & 0.7594 & 0.9277 & -- \\
& Greedy (G) & 0.9305 & 0.3707 & 0.9750 & 4.15 & 0.9152 & 0.4841 & 0.9484 & 4.25 \\
& PhaseWin (PW) & 0.9253 & 0.4159 & 0.9735 & 2.24 & 0.9090 & 0.5254 & 0.9466 & 1.54 \\
& \bluecell{\textsc{CoPAIR}+PW} & \bluecell{0.9308} & \bluecell{0.4188} & \bluecell{0.9742} & \bluecell{\(2.21_{-.03}\)} & \bluecell{0.9104} & \bluecell{0.5226} & \bluecell{0.9466} & \bluecell{\(1.54_{0.00}\)} \\
& \bluecell{\textsc{TRACE}+PW} & \bluecell{0.9299} & \bluecell{0.4213} & \bluecell{0.9748} & \bluecell{\(2.10_{-.14}\)} & \bluecell{0.9102} & \bluecell{0.5216} & \bluecell{0.9481} & \bluecell{\(1.43_{-.11}\)} \\
& \bluecell{\textsc{CoPAIR}+G} & \bluecell{0.9347} & \bluecell{0.3693} & \bluecell{0.9763} & \bluecell{\(4.09_{-.05}\)} & \bluecell{0.9161} & \bluecell{0.4738} & \bluecell{0.9495} & \bluecell{\(4.22_{-.04}\)} \\
& \bluecell{\textsc{TRACE}+G} & \bestcell{0.9353} & \bluecell{0.3674} & \bluecell{0.9758} & \bluecell{\(3.98_{-.16}\)} & \bestcell{0.9165} & \bluecell{0.4693} & \bluecell{0.9499} & \bluecell{\(4.10_{-.15}\)} \\
\midrule
\multirow{8}{*}{RePOPE}
& LLaVA-CAM & 0.3745 & 0.3919 & 0.6873 & -- & 0.3697 & 0.3628 & 0.6737 & -- \\
& IGOS++ & 0.3721 & 0.3874 & 0.6916 & -- & 0.3592 & 0.3784 & 0.6815 & -- \\
& Greedy (G) & 0.8055 & 0.1284 & 0.9245 & 4.09 & 0.7683 & 0.1603 & 0.8982 & 4.24 \\
& PhaseWin (PW) & 0.7783 & 0.1424 & 0.9122 & 2.06 & 0.7487 & 0.1719 & 0.8893 & 1.71 \\
& \bluecell{\textsc{CoPAIR}+PW} & \bluecell{0.7820} & \bluecell{0.1458} & \bluecell{0.9148} & \bluecell{\(2.05_{-.01}\)} & \bluecell{0.7553} & \bluecell{0.1719} & \bluecell{0.8925} & \bluecell{\(1.64_{-.07}\)} \\
& \bluecell{\textsc{TRACE}+PW} & \bluecell{0.7810} & \bluecell{0.1497} & \bluecell{0.9179} & \bluecell{\(1.93_{-.13}\)} & \bluecell{0.7620} & \bluecell{0.1721} & \bluecell{0.8959} & \bluecell{\(1.46_{-.25}\)} \\
& \bluecell{\textsc{CoPAIR}+G} & \bluecell{0.8028} & \bluecell{0.1302} & \bluecell{0.9229} & \bluecell{\(4.01_{-.08}\)} & \bluecell{0.7779} & \bluecell{0.1606} & \bluecell{0.9042} & \bluecell{\(4.16_{-.08}\)} \\
& \bluecell{\textsc{TRACE}+G} & \bluecell{0.8045} & \bluecell{0.1322} & \bestcell{0.9273} & \bluecell{\(3.88_{-.21}\)} & \bluecell{0.7859} & \bluecell{0.1589} & \bestcell{0.9082} & \bluecell{\(3.99_{-.25}\)} \\
\bottomrule
\end{tabular}}
\vspace{-10pt}
\end{table}
\FloatBarrier

\subsection{Direct Evidence Masks for RePOPE Repair}
\label{subsec:direct-mask-repair}

The previous tables evaluate ordered insertion curves. Table~\ref{tab:direct-repair-main}
tests the mask-first use case directly: a single selected evidence mask is applied as
one repair intervention on RePOPE. This setting does not require ranking every region.
Direct \textsc{TRACE} masks repair 94.44\% of Qwen cases and 96.00\% of LLaVA cases,
substantially above the LLaVA-CAM and IGOS++ full-curve repair rates. \textsc{CoPAIR}
Direct is also stronger than these non-search baselines, but the gap between
\textsc{CoPAIR} and \textsc{TRACE} is large, matching our method design: coarse
pairing is a useful initializer, whereas direct fine-region mask search is the more
appropriate tool when the output itself is a compact evidence mask. The full ordered
repair protocol, including CSR@10, is reported in Appendix~\ref{app:eagle-repair-results};
the memory-saving deployment used for the parenthesized engineered memory numbers in Table~\ref{tab:direct-repair-main}
is detailed in Appendix~\ref{app:memory-saving-settings}. Figure~\ref{fig:repope-repair-vis}
visualizes representative cases where full-curve LLaVA-CAM and IGOS++ do not find a
repairing prefix, while \textsc{TRACE} Direct returns a compact mask that repairs the
model answer.

\begin{table}[t!]
\centering
\vspace{-14pt}
\caption{\textbf{Single-point RePOPE repair using direct evidence masks.} LLaVA-CAM and IGOS++ rows are full-curve baselines from the repair protocol; \textsc{CoPAIR} Direct and \textsc{TRACE} Direct apply one compact mask. Gold cells mark the best single-point repair result for each model. Memory and GPU time are averaged over 100 RePOPE adversarial samples; for LLaVA-CAM and IGOS++, memory is reported as original (engineered).}
\label{tab:direct-repair-main}
\scriptsize
\setlength{\tabcolsep}{3.4pt}
\renewcommand{\arraystretch}{0.96}
\resizebox{0.8\linewidth}{!}{
\begin{tabular}{@{}clcccccc@{}}
\toprule
Model & Method & Protocol & \primaryhead{Repair Rate} & \primaryhead{Correct/Total} & \primaryhead{Failed} & Mem. (GB) & Time (s) \\
\midrule
\multirow{4}{*}{Qwen2.5-VL-3B} & LLaVA-CAM & full curve & 69.63 & 188/270 & 82 & 27.1 (16.0) & 11.3 \\
& IGOS++ & full curve & 72.59 & 196/270 & 74 & 35.2 (18.4) & 22.0 \\
& \bluecell{\textsc{CoPAIR} Direct} & \bluecell{single point} & \bluecell{84.44} & \bluecell{228/270} & \bluecell{42} & \bluecell{7.2} & \bluecell{12.7} \\
& \bluecell{\textsc{TRACE} Direct} & \bluecell{single point} & \bestcell{94.44} & \bestcell{255/270} & \bestcell{15} & \bluecell{7.2} & \bluecell{15.8} \\
\midrule
\multirow{4}{*}{LLaVA-v1.5-7B} & LLaVA-CAM & full curve & 68.33 & 205/300 & 95 & 37.1 (16.3) & 15.4 \\
& IGOS++ & full curve & 70.33 & 211/300 & 89 & 93.9 (18.9) & 22.5 \\
& \bluecell{\textsc{CoPAIR} Direct} & \bluecell{single point} & \bluecell{82.67} & \bluecell{248/300} & \bluecell{52} & \bluecell{13.4} & \bluecell{17.0} \\
& \bluecell{\textsc{TRACE} Direct} & \bluecell{single point} & \bestcell{96.00} & \bestcell{288/300} & \bestcell{12} & \bluecell{13.4} & \bluecell{22.3} \\
\bottomrule
\end{tabular}}
\vspace{-10pt}
\end{table}

\begin{figure}[t]
\centering
\includegraphics[width=0.8\linewidth]{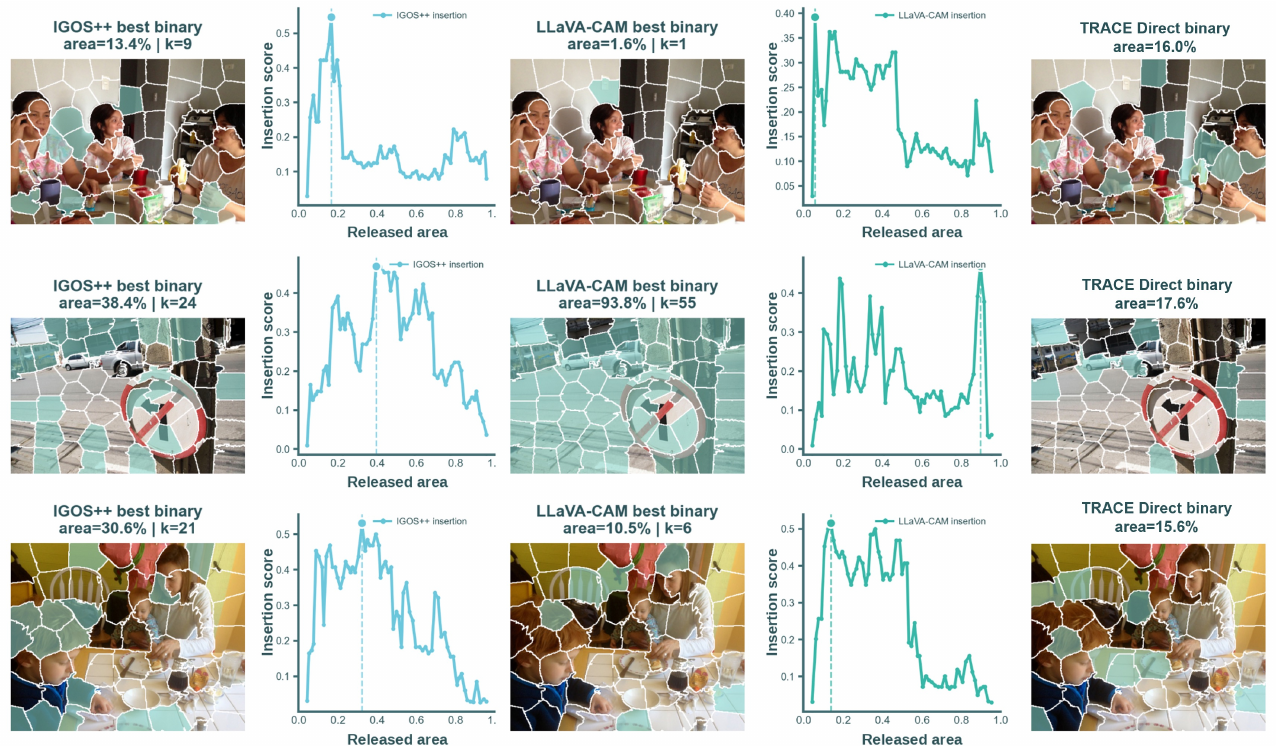}
\caption{\textbf{Qualitative RePOPE repair examples.} For each example, the full-curve LLaVA-CAM and
IGOS++ baselines are visualized with their best binary masks and insertion curves, but
the swept curves do not yield a repairing prefix. \textsc{TRACE} Direct instead returns
one compact evidence mask that repairs the hallucinated answer.}
\label{fig:repope-repair-vis}
\vspace{-14pt}
\end{figure}
\FloatBarrier

\section{Conclusion}
\label{sec:conclusion}

We presented \textsc{TRACE}, a forward-only method for faithful top-\(k\) visual
attribution. It directly searches fixed-size fine-region evidence masks and delivers
two operating modes: a standalone binary-mask mode for tasks that require only a compact
evidence set, and an initialization mode that improves full-ordering Greedy or PhaseWin
search.
\textsc{CoPAIR} provides a complementary deterministic initializer based on coarse
spatial grouping and pairwise evidence enumeration, but the central contribution is the
\textsc{TRACE} mask: it handles non-geometric fine-region combinations, can support
repair directly, and reduces redundant early search when a complete ranking is required.
Both are model-agnostic and validated on image classification (ImageNet-1K, SLICO-64;
CLIP ViT-L/14, CLIP RN101, ResNet-101), MLLM decision-faithfulness attribution on
POPE, and repair-oriented evidence recovery on RePOPE (Qwen2.5-VL-3B-Instruct and
LLaVA-v1.5-7B).

\paragraph{Limitations.}
The initialization overhead ($c^2$ for \textsc{CoPAIR} or $RM$ for \textsc{TRACE}) is
negligible for large $n$ but may dominate when $n$ is small. \textsc{TRACE}'s stochastic
sampling introduces run-to-run variance, which is small in practice but non-zero. The
fixed-\(k\) constraint in \textsc{TRACE} requires choosing mask size \(k\) in advance.




\bibliographystyle{iclr2026_conference}
\bibliography{references}


\appendix
\section{Additional TRACE Notes}
\label{app:trace-details}

Section~\ref{subsec:trace} gives the \textsc{TRACE} objective, Gumbel-top-\(k\)
sampling rule, elite update, and continuation interface. This appendix avoids
restating those equations and records only the implementation-level conventions and
scope limitations that are useful when interpreting the experiments.

\subsection{Candidate Masks and Returned Objects}
\label{app:trace-candidate-records}

\textsc{TRACE} samples only fixed-cardinality fine-region masks. Every evaluated
candidate therefore contains exactly \(k\) distinct regions, and no rejection step or
post-hoc cardinality repair is needed. In the implementation, a candidate mask is
identified by the sorted set of selected region indices. This representation is used
for duplicate detection and for storing the best observed candidate during the
cross-entropy rounds.

The returned mask is empirical rather than certified: it is the highest-scoring
candidate observed under the allocated model-evaluation budget. This distinction
matters because the proxy \(F(S)\) is a black-box, nonconvex set objective induced by
masked model responses. We therefore do not claim that \textsc{TRACE} solves the
global optimum over all \(\binom{n}{k}\) masks; the method is a budgeted search
procedure that concentrates samples around regions repeatedly appearing in elite
fixed-size masks.

When the downstream task is mask-oriented, this best observed set is returned directly
as the attribution or repair mask. When the downstream task requires a full ordering,
the same set is used as the initialized prefix described in
Eq.~\eqref{eq:initialized-search}. Thus the direct-mask and full-ordering uses share
one searched object, rather than using separate attribution procedures.

\subsection{Why Not Independent Region Scores?}
\label{app:trace-score-rationale}

A one-step score for each region is cheaper, but it does not model the interaction
among selected regions. In visual attribution, this interaction is common: two object
parts may be redundant, a background fragment may only matter together with an object
region, and an MLLM answer token may depend on a specific combination of local visual
cues. Independent scores also do not enforce the output contract that many downstream
uses require: a compact mask of exactly \(k\) regions.

Independent Bernoulli sampling has the opposite issue. It can model uncertainty over
regions, but the sampled cardinality varies unless an extra constraint or projection is
introduced. \textsc{TRACE} uses Gumbel-top-\(k\) sampling because the sampled object is
already feasible for the target mask family. This keeps the optimization distribution,
the evaluated candidates, and the final output in the same fixed-cardinality space.

\subsection{Budget and Caching Scope}
\label{app:trace-budget-scope}

The nominal \textsc{TRACE} budget is \(RM\) candidate masks for \(R\) rounds and \(M\)
samples per round. Duplicate masks can occur, especially after the sampling
distribution becomes concentrated. In that case, the implementation reuses the stored
score instead of issuing another model call. The MEC values reported for initialized
rows are inclusive: they include the initializer calls that survive caching and the
subsequent continuation calls made by Greedy or PhaseWin.

This accounting is deliberately narrower than wall-clock runtime. MEC is a
forward-call count for comparable forward-only selection procedures; it does not
attempt to price backward passes, optimizer steps, KV-cache memory, or hardware
placement. Those runtime and memory details are reported separately where they matter,
especially for the MLLM repair comparison in Table~\ref{tab:direct-repair-main}.

\subsection{Probabilistic Recovery Under a Phase-Structured Attribution Model}
\label{app:trace-recovery-guarantee}

We next record a finite-budget recovery bound for \textsc{TRACE}. The structural
assumptions used in this subsection are not new modeling claims of this paper. They
are inherited from the extended PhaseWin analysis~\citep{gu2026phasewinExtended},
which interprets the hard tail of search-based visual attribution through
partition-sensitive semantic evidence blocks. This inheritance includes the
block-level monotonicity assumption used below: adding activated semantic evidence
blocks should not decrease the dominant attribution value. We use the same
phase-structured view only to make explicit when the cross-entropy updates in
\textsc{TRACE} can improve over non-adaptive random search.

Let \(U=\{u_1,\ldots,u_n\}\) be the fine-region set and let
\begin{equation}
    \mathcal{C}_k(U)=\{S\subseteq U: |S|=k\}
\end{equation}
be the feasible family of exactly-\(k\) masks. For a nonnegative attribution proxy
\(F:2^U\to\mathbb{R}_+\), define the fixed-\(k\) optimum and the best
\textsc{TRACE} value observed under a budget of \(R\) rounds and
\(M_{\mathrm{samp}}\) masks per round as
\begin{equation}
    \mathrm{opt}
    =
    \max_{S\in\mathcal{C}_k(U)}F(S),
    \qquad
    \mathrm{opt}_{\mathrm{trace}}
    =
    \max_{\substack{0\le r\le R-1\\1\le j\le M_{\mathrm{samp}}}}
    F(S_{r,j}),
    \label{eq:app-trace-opt-values}
\end{equation}
where \(\mathrm{opt}>0\). We write \(M_{\mathrm{samp}}\) in this subsection to
avoid confusing the per-round sample count with the optimum value; it is the same
quantity denoted by \(M\) in Algorithm~\ref{alg:trace}. For an approximation ratio
\(\mu\in(0,1)\), define the near-optimal mask set
\begin{equation}
    \mathcal{G}_\mu
    =
    \{S\in\mathcal{C}_k(U): F(S)\ge \mu\,\mathrm{opt}\}.
    \label{eq:app-trace-near-optimal-set}
\end{equation}
Then \(\mathrm{opt}_{\mathrm{trace}}/\mathrm{opt}\ge \mu\) holds exactly when at
least one evaluated mask lies in \(\mathcal{G}_\mu\).

Let \(Q_{\theta_r}\) be the Gumbel-top-\(k\) sampling distribution used by
\textsc{TRACE} in round \(r\), and define
\begin{equation}
    p_r^\mu
    =
    Q_{\theta_r}(\mathcal{G}_\mu).
    \label{eq:app-trace-round-hit-prob}
\end{equation}
Conditioned on the round-\(r\) sampling distribution, the probability that all
\(M_{\mathrm{samp}}\) masks miss \(\mathcal{G}_\mu\) is
\((1-p_r^\mu)^{M_{\mathrm{samp}}}\). Therefore,
\begin{equation}
    \Pr\left[
    \frac{\mathrm{opt}_{\mathrm{trace}}}{\mathrm{opt}}
    \ge \mu
    \right]
    \ge
    1-
    \prod_{r=0}^{R-1}(1-p_r^\mu)^{M_{\mathrm{samp}}}.
    \label{eq:app-trace-generic-hit-bound}
\end{equation}
This identity is correct but hides the attribution geometry inside \(p_r^\mu\). The
PhaseWin-structured model makes this term more interpretable.

Assume, following the extended PhaseWin abstraction, that the fine regions admit a
semantic block partition
\begin{equation}
    \mathcal{H}=\{H_1,\ldots,H_q\},
    \qquad
    H_a\cap H_b=\emptyset\ (a\ne b),
    \qquad
    \bigcup_{j=1}^q H_j=U .
\end{equation}
For a mask \(S\), let
\begin{equation}
    \mathsf{B}(S)
    =
    \{j\in[q]: S\cap H_j\ne\emptyset\}
\end{equation}
be its activated block set. Following the PhaseWin extended model, the proxy has a
dominant monotone block-level component plus a bounded nonnegative residual:
\begin{equation}
    F(S)=\Phi(\mathsf{B}(S))+\Delta(S),
    \qquad
    0\le \Delta(S)\le |S|\epsilon_\Delta,
    \label{eq:app-phase-decomposition}
\end{equation}
where the monotonicity of \(\Phi\), namely
\(J\subseteq J'\Rightarrow \Phi(J)\le\Phi(J')\), is the PhaseWin block-evidence
monotonicity assumption rather than an additional assumption introduced by
\textsc{TRACE}. Define the best block-level value attainable by at most \(k\) blocks:
\begin{equation}
    \Phi_k^\star
    =
    \max_{J\subseteq[q],\, |J|\le k}\Phi(J).
\end{equation}
Because an exactly-\(k\) mask activates at most \(k\) blocks,
\begin{equation}
    \mathrm{opt}
    \le
    \Phi_k^\star+k\epsilon_\Delta .
    \label{eq:app-opt-upper-phase}
\end{equation}

Define the block-certificate family
\begin{equation}
    \mathcal{A}_\mu
    =
    \left\{
    J\subseteq[q]:
    |J|\le k,\ 
    \Phi(J)\ge
    \mu(\Phi_k^\star+k\epsilon_\Delta)
    \right\}.
    \label{eq:app-block-certificate-family}
\end{equation}
Every \(J\in\mathcal{A}_\mu\) certifies a \(\mu\)-near-optimal basin: if
\(\mathsf{B}(S)\supseteq J\), then by monotonicity and
Eq.~\eqref{eq:app-opt-upper-phase},
\begin{equation}
    F(S)
    \ge
    \Phi(\mathsf{B}(S))
    \ge
    \Phi(J)
    \ge
    \mu(\Phi_k^\star+k\epsilon_\Delta)
    \ge
    \mu\,\mathrm{opt}.
\end{equation}
Hence
\begin{equation}
    \{S: \mathsf{B}(S)\supseteq J\}
    \subseteq
    \mathcal{G}_\mu .
    \label{eq:app-certificate-implies-good}
\end{equation}
This converts exact-mask recovery into semantic basin recovery.

Under the equivalent Plackett--Luce view of Gumbel-top-\(k\) sampling, let
\begin{equation}
    w_{r,i}=\exp(\theta_{r,i}/T),
    \qquad
    W_r=\sum_{i\in U}w_{r,i},
    \qquad
    \eta_{r,j}
    =
    \frac{\sum_{i\in H_j}w_{r,i}}{W_r}
    \label{eq:app-block-mass}
\end{equation}
be the normalized mass assigned to semantic block \(H_j\) in round \(r\). For a
certificate \(J\in\mathcal{A}_\mu\) with \(h=|J|\), a sufficient event for
activating all blocks in \(J\) is that the first \(h\) sampled positions hit the
\(h\) blocks in \(J\) in some order. Thus,
\begin{equation}
    Q_{\theta_r}\bigl(\mathsf{B}(S)\supseteq J\bigr)
    \ge
    h!\prod_{j\in J}\eta_{r,j}.
    \label{eq:app-prefix-block-bound}
\end{equation}
The events in Eq.~\eqref{eq:app-prefix-block-bound} are disjoint across
certificates of the same cardinality. A complementary all-\(k\) lower bound follows
from a union bound over missing blocks:
\begin{equation}
    Q_{\theta_r}\bigl(\mathsf{B}(S)\supseteq J\bigr)
    \ge
    \left[
    1-\sum_{j\in J}(1-\eta_{r,j})^k
    \right]_+,
    \qquad
    [x]_+=\max\{x,0\}.
    \label{eq:app-union-block-bound}
\end{equation}
Let \(\mathcal{A}_{\mu,h}=\{J\in\mathcal{A}_\mu: |J|=h\}\). Combining
Eqs.~\eqref{eq:app-certificate-implies-good}--\eqref{eq:app-union-block-bound},
define
\begin{equation}
    \underline p_r^\mu
    =
    \max
    \left\{
    \max_{1\le h\le k}
    \sum_{J\in\mathcal{A}_{\mu,h}}
    h!\prod_{j\in J}\eta_{r,j},
    \ 
    \max_{J\in\mathcal{A}_\mu}
    \left[
    1-\sum_{j\in J}(1-\eta_{r,j})^k
    \right]_+
    \right\}.
    \label{eq:app-structured-pr-lower}
\end{equation}
Then
\begin{equation}
    Q_{\theta_r}(\mathcal{G}_\mu)
    \ge
    \underline p_r^\mu.
\end{equation}
Substitution into Eq.~\eqref{eq:app-trace-generic-hit-bound} gives the
phase-structured recovery guarantee:
\begin{equation}
    \Pr\left[
    \frac{\mathrm{opt}_{\mathrm{trace}}}{\mathrm{opt}}
    \ge \mu
    \right]
    \ge
    1-
    \prod_{r=0}^{R-1}(1-\underline p_r^\mu)^{M_{\mathrm{samp}}}.
    \label{eq:app-phase-structured-guarantee}
\end{equation}
This bound is more informative than a uniform exact-mask bound because it depends on
semantic block-basin mass rather than only on the fraction of good masks inside the
full \(\binom{n}{k}\) combinatorial space.

The cross-entropy update enters by increasing the block-basin mass across rounds. For
a certificate \(J\in\mathcal{A}_\mu\), define
\begin{equation}
    \eta_{r,J}=\sum_{j\in J}\eta_{r,j}.
\end{equation}
Assume the update amplifies the odds of this basin:
\begin{equation}
    \frac{\eta_{r+1,J}}{1-\eta_{r+1,J}}
    \ge
    \lambda_{r,J}
    \frac{\eta_{r,J}}{1-\eta_{r,J}},
    \qquad
    \lambda_{r,J}\ge 1.
    \label{eq:app-odds-amplification}
\end{equation}
With
\begin{equation}
    \Lambda_{r,J}=\prod_{\ell=0}^{r-1}\lambda_{\ell,J},
    \qquad
    \Lambda_{0,J}=1,
\end{equation}
we obtain
\begin{equation}
    \eta_{r,J}
    \ge
    z_{r,J}
    :=
    \frac{\Lambda_{r,J}\eta_{0,J}}
    {1-\eta_{0,J}+\Lambda_{r,J}\eta_{0,J}}.
    \label{eq:app-z-basin-mass}
\end{equation}
If the mass inside a certificate is not collapsed onto a single block, namely
\begin{equation}
    \eta_{r,j}\ge b_J\eta_{r,J},
    \qquad
    \forall j\in J,
    \qquad
    0<b_J\le |J|^{-1},
    \label{eq:app-internal-balance}
\end{equation}
then, for \(h_J=|J|\),
\begin{equation}
    Q_{\theta_r}(\mathcal{G}_\mu)
    \ge
    \max
    \left\{
    h_J! b_J^{h_J} z_{r,J}^{h_J},
    \left[
    1-h_J(1-b_J z_{r,J})^k
    \right]_+
    \right\}.
    \label{eq:app-balanced-certificate-bound}
\end{equation}
Consequently,
\begin{equation}
    \begin{aligned}
    \Pr\left[
    \frac{\mathrm{opt}_{\mathrm{trace}}}{\mathrm{opt}}
    \ge \mu
    \right]
    \ge
    1-
    \prod_{r=0}^{R-1}
    \Bigg(
    1
    -
    \max_{J\in\mathcal{A}_\mu}
    \max
    \Big\{
    h_J! b_J^{h_J} z_{r,J}^{h_J},
    \left[
    1-h_J(1-b_J z_{r,J})^k
    \right]_+
    \Big\}
    \Bigg)^{M_{\mathrm{samp}}}.
    \end{aligned}
    \label{eq:app-ce-aware-guarantee}
\end{equation}
In the homogeneous case \(\lambda_{r,J}=\lambda>1\),
\begin{equation}
    z_{r,J}
    =
    \frac{\lambda^r\eta_{0,J}}
    {1-\eta_{0,J}+\lambda^r\eta_{0,J}}.
    \label{eq:app-homogeneous-z}
\end{equation}
Thus, under the PhaseWin-style block-basin model, the cross-entropy update improves
the recovery probability by increasing the probability mass assigned to
near-optimal semantic basins over successive rounds.

\paragraph{Numerical estimate.}
For the setting used in the estimate, take \(R=5\), \(M_{\mathrm{samp}}=32\),
total budget \(B_{\mathrm{eval}}=160\), and \(\mu=0.8\). If no cross-entropy
improvement is assumed and \(p_r^{0.8}=p_0\) is fixed, then
\begin{equation}
    P_{\mathrm{hit}}
    =
    1-(1-p_0)^{160}.
\end{equation}
Representative values are
\begin{center}
\small
\begin{tabular}{cc}
\toprule
\(p_0\) & \(P_{\mathrm{hit}}\) \\
\midrule
\(0.2\%\) & \(27.4\%\) \\
\(0.5\%\) & \(55.2\%\) \\
\(1.0\%\) & \(80.0\%\) \\
\(2.0\%\) & \(96.1\%\) \\
\bottomrule
\end{tabular}
\end{center}
With homogeneous cross-entropy odds amplification,
\begin{equation}
    p_r
    =
    \frac{\lambda^r p_0}
    {1-p_0+\lambda^r p_0},
    \qquad
    P_{\mathrm{hit}}
    =
    1-\prod_{r=0}^{4}(1-p_r)^{32}.
\end{equation}
For \(\lambda=1.5\), the same initial probabilities yield
\begin{center}
\small
\begin{tabular}{cc}
\toprule
\(p_0\) & \(P_{\mathrm{hit}}\) \\
\midrule
\(0.2\%\) & \(56.9\%\) \\
\(0.5\%\) & \(87.8\%\) \\
\(1.0\%\) & \(98.5\%\) \\
\(2.0\%\) & \(99.98\%\) \\
\bottomrule
\end{tabular}
\end{center}
For a block-basin interpretation, suppose the \(\mu=0.8\) certificate consists of
\(h\) semantic blocks, its initial basin mass is \(\eta_0\), the amplification is
\(\lambda=1.5\), and the certificate mass is approximately balanced across its
blocks, \(b\approx 1/h\). Using the simplified lower bound
\begin{equation}
    h_r
    \approx
    h!b^h z_r^h,
    \qquad
    z_r=
    \frac{\lambda^r\eta_0}
    {1-\eta_0+\lambda^r\eta_0},
    \qquad
    P_{\mathrm{hit}}
    \approx
    1-\prod_{r=0}^{4}(1-h_r)^{32},
\end{equation}
we obtain
\begin{center}
\small
\begin{tabular}{ccc}
\toprule
\(h\) & \(\eta_0\) & \(P_{\mathrm{hit}}\) \\
\midrule
\(1\) & \(5\%\) & \(\approx 100\%\) \\
\(2\) & \(5\%\) & \(75.0\%\) \\
\(2\) & \(10\%\) & \(98.9\%\) \\
\(3\) & \(5\%\) & \(9.8\%\) \\
\(3\) & \(10\%\) & \(43.0\%\) \\
\(3\) & \(20\%\) & \(91.7\%\) \\
\bottomrule
\end{tabular}
\end{center}
These estimates suggest that, under moderate cross-entropy amplification, five
rounds with 32 samples per round are sufficient to obtain high recovery probability
when the \(0.8\)-near-optimal basin is controlled by one or two dominant semantic
blocks. A conservative operating estimate,
\(p_0^{0.8}=0.5\%\) and \(\lambda=1.5\), gives
\(P_{\mathrm{hit}}\approx 87.8\%\). A more favorable estimate,
\(p_0^{0.8}=1.0\%\) and \(\lambda=1.5\), gives
\(P_{\mathrm{hit}}\approx 98.5\%\). Without adaptive amplification, the same budget
gives \(P_{\mathrm{hit}}\approx 55\%\)--\(80\%\) for
\(p_0^{0.8}\in[0.5\%,1.0\%]\).

\subsection{Notation for the Recovery Bound}
\label{app:trace-recovery-notation}

\begin{table}[H]
\centering
\small
\begin{tabular}{@{}p{0.18\linewidth}p{0.22\linewidth}p{0.50\linewidth}@{}}
\toprule
Symbol & Type & Meaning \\
\midrule
\(U=\{u_i\}_{i=1}^n\) & fine regions & Image regions used by the attribution search. \\
\(\mathcal{C}_k(U)\) & feasible mask family & All exactly-\(k\) subsets of \(U\). \\
\(F(S)\) & proxy score & Nonnegative attribution objective evaluated on mask \(S\). \\
\(\mathrm{opt}\) & scalar & Exhaustive fixed-\(k\) optimum over \(\mathcal{C}_k(U)\). \\
\(\mathrm{opt}_{\mathrm{trace}}\) & scalar & Best value observed by \textsc{TRACE} under the finite budget. \\
\(\mu\) & ratio & Target approximation threshold in \((0,1)\). \\
\(\mathcal{G}_\mu\) & mask set & Masks with \(F(S)\ge \mu\,\mathrm{opt}\). \\
\(R\) & integer & Number of \textsc{TRACE} cross-entropy rounds. \\
\(M_{\mathrm{samp}}\) & integer & Masks sampled per round; this corresponds to \(M\) in Algorithm~\ref{alg:trace}. \\
\(Q_{\theta_r}\) & distribution & Round-\(r\) Gumbel-top-\(k\) sampling distribution. \\
\(p_r^\mu\) & probability & \(Q_{\theta_r}(\mathcal{G}_\mu)\), the round-\(r\) near-optimal hit probability. \\
\(\mathcal{H}=\{H_j\}_{j=1}^q\) & semantic partition & PhaseWin-style partition of fine regions into semantic evidence blocks. \\
\(\mathsf{B}(S)\) & block set & Indices of semantic blocks activated by mask \(S\). \\
\(\Phi\) & set function & Monotone block-level dominant attribution component. \\
\(\Delta(S)\) & residual & Bounded nonnegative fine-region residual in the block model. \\
\(\epsilon_\Delta\) & scalar & Per-region upper bound for the residual contribution. \\
\(\Phi_k^\star\) & scalar & Best block-level value over at most \(k\) semantic blocks. \\
\(\mathcal{A}_\mu\) & certificate family & Block sets certifying \(\mu\)-near-optimality. \\
\(w_{r,i}\) & positive weight & \(\exp(\theta_{r,i}/T)\), the round-\(r\) sampling weight of region \(u_i\). \\
\(\eta_{r,j}\) & probability mass & Normalized round-\(r\) mass assigned to block \(H_j\). \\
\(\underline p_r^\mu\) & probability lower bound & Phase-structured lower bound on \(p_r^\mu\). \\
\(\eta_{r,J}\) & probability mass & Total round-\(r\) sampling mass of certificate \(J\). \\
\(\lambda_{r,J}\) & amplification factor & Cross-entropy odds amplification for certificate \(J\) from round \(r\) to \(r+1\). \\
\(\Lambda_{r,J}\) & cumulative factor & Product of amplification factors through round \(r\). \\
\(z_{r,J}\) & probability mass lower bound & Lower bound on \(\eta_{r,J}\) after cumulative amplification. \\
\(b_J\) & balance factor & Lower bound on each block's share inside certificate \(J\). \\
\bottomrule
\end{tabular}
\caption{Notation used in the phase-structured recovery bound for \textsc{TRACE}.}
\label{tab:trace-recovery-notation}
\end{table}

\section{Extended Attribution Background}
\label{app:attribution-background}

This appendix expands the compact Related Work in Section~\ref{sec:related}. The goal
is not to provide another short list of baselines, but to make explicit the historical
shift behind our problem formulation: visual attribution has moved from producing
visually plausible heatmaps, to testing evidence by perturbation, to modeling
region-level coalitions and ordered subsets under the cost constraints of modern
vision-language models. Following the terminology used throughout this paper, we refer
to the LIMA/VPS/PhaseWin line as \emph{ordered subset search}: these methods construct
an ordering of image regions so that its prefixes can be replayed by insertion or
deletion.

\subsection{From Heatmaps to Evidence Search}
\label{app:history-from-heatmaps}

The earliest visual-attribution interface is a saliency map: each pixel or region is
assigned an importance score. This interface is intuitive, but a heatmap alone does not
answer whether the highlighted evidence is sufficient for the model output. Faithful
attribution therefore adds a behavioral test. If the most important regions are
inserted first, the target response should recover quickly; if they are deleted first,
the target response should drop quickly. RISE~\citep{petsiuk2018rise} made this
insertion--deletion evaluation protocol central to black-box visual explanation, and
the same operational view underlies later search-based methods.

The combinatorial nature of the problem is immediate. A pixel-level mask over \(N\)
pixels has \(2^N\) possible retained subsets. Practical methods reduce this to a
region set \(U=\{u_1,\ldots,u_n\}\) defined by superpixels, patches, proposals, or
semantic masks, but the retained-subset family still has size \(2^n\). This reduction
creates the main design axis in modern faithful attribution: should we infer
independent region scores, estimate average coalition contribution, optimize a
continuous mask, or search directly for high-value region subsets and prefixes?

\subsection{Gradient and Activation Saliency}
\label{app:gradient-saliency-history}

Gradient saliency starts from local sensitivity. Simonyan et al.~\citep{simonyan2014deep}
computed the target class score gradient with respect to the input image and used it as
an image-specific saliency map. Zeiler and Fergus~\citep{zeiler2014visualizing}
developed DeconvNet visualizations and occlusion sensitivity to inspect CNN feature
hierarchies from the complementary perspective of internal activation and localized
input removal. Layer-wise Relevance Propagation~\citep{bach2015pixel} then framed
explanation as redistributing the prediction score backward through the network to
obtain pixel-level relevance. Guided Backpropagation, introduced through the all
convolutional network analysis of Springenberg et al.~\citep{springenberg2015striving},
modified the backward ReLU signal to produce sharper edge-like maps. SmoothGrad
\citep{smilkov2017smoothgrad} reduced visual noise by averaging saliency maps over
noisy input copies, while Integrated Gradients~\citep{sundararajan2017axiomatic}
introduced an axiomatic path-integral attribution from a baseline input to the observed
input.

Activation-map methods moved the saliency interface from input gradients to spatial
feature maps. CAM~\citep{zhou2016learning} showed that CNNs with global average pooling
can localize class-discriminative image regions from image-level supervision. Grad-CAM
\citep{selvaraju2020grad} generalized this idea by weighting convolutional feature maps
with target gradients, making class-discriminative localization applicable to a broader
range of architectures and tasks. Grad-CAM++~\citep{chattopadhyay2018gradcampp}
improved multi-instance localization through a weighted combination of positive
gradients, and Score-CAM~\citep{wang2020score} replaced gradient weights with
forward-score weights over activation-map masks, partly bridging activation maps and
perturbation testing.

The same gradient/activation logic was later adapted to new architectures and outputs.
ViT-CX~\citep{xie2023vitcx} argued that attention weights over ViT patches are not
enough and instead modeled the causal effect of patch embeddings. Grad-ECLIP
\citep{zhao2024gradient} adapted gradient explanation to transformer-based CLIP
\citep{radford2021learning}. For object detectors, Spatial Sensitive Grad-CAM++
\citep{yamauchi2024spatial} extended Grad-CAM++ style weighting to detector-specific
heatmaps, while ODAM~\citep{zhao2024odam} produced instance-specific detector
activation maps from gradients of detection targets. For MLLMs, TAM~\citep{li2025tam}
further changed the explanation unit from a single class score to generated tokens,
explicitly addressing context interference among autoregressive tokens.

This line solves the speed problem. With white-box access, a small number of backward
passes can produce a heatmap, making gradient and activation saliency useful for rapid
inspection and model debugging. Its limitation is that local sensitivity and internal
activation are not the same as evidence sufficiency. The sanity checks of Adebayo et
al.~\citep{adebayo2018sanity} showed that visually appealing saliency maps can fail to
reflect model parameters or data dependence. This is precisely why faithful attribution
requires perturbation replay or subset search when the goal is to test what the model
actually relies on.

\subsection{Perturbation Testing and Black-Box Attribution}
\label{app:perturbation-history}

Perturbation methods treat attribution as an input-output experiment. LIME
\citep{ribeiro2016why} samples local perturbations around an input, fits an
interpretable surrogate model, and uses the surrogate coefficients as an explanation;
in images, the interpretable units are usually superpixels. Meaningful Perturbation
\citep{fong2017interpretable} optimized a mask that removes or blurs a small image
region while reducing the target response, making explanation explicitly testable by
editing the input. Extremal Perturbations~\citep{fong2019understanding} refined this
idea with fixed-area constraints and smooth masks, turning perturbation analysis into a
more controlled optimization problem.

RISE~\citep{petsiuk2018rise} took a simpler black-box route: sample many random binary
masks, evaluate the target score under each masked image, and aggregate the masks
weighted by their scores. D-RISE~\citep{petsiuk2021black} extended the same idea to
object detectors by defining a detection-similarity score that combines classification
and localization behavior. HSIC Attribution~\citep{novello2022making} then used a
dependence measure to estimate input-output relationships more efficiently than
plain random-mask averaging.

These methods are closer to faithful attribution than pure gradients because they
measure model behavior under retained or removed evidence. They also make black-box
explanation possible. Their remaining weakness is the output contract: random masks,
optimized continuous masks, or local surrogate coefficients usually collapse into a
per-region saliency score. That score can be replayed by insertion or deletion, but it
was not necessarily learned to optimize the insertion prefix itself.

\subsection{Shapley-Style Region Contribution}
\label{app:shapley-history}

Shapley value~\citep{shapley1953value} provides a principled way to assign average
coalition contribution to players. In visual attribution, the players can be
superpixels, patches, object proposals, or semantic masks, and the coalition value is a
model score \(G(S)\) under a retained region set \(S\). SHAP
\citep{lundberg2017unified} made Shapley-style additive feature attribution a standard
model-explanation framework. Compared with one-region occlusion or local gradients,
this view explicitly recognizes that an image region may matter differently depending
on which other regions are present.

The cost and semantics of Shapley-style explanation are also important. Exact Shapley
values require averaging over exponentially many coalitions, so image applications rely
on sampling or approximations. Kumar et al.~\citep{kumar2020problems} further argued
that Shapley-based feature importance has nontrivial mathematical and semantic limits,
especially when feature dependence and causal interpretation matter. Nevertheless,
Shapley-style ideas have shaped region-level visual diagnosis: Explain to Fix
\citep{gudovskiy2018explain} used approximate Shapley feature importance to interpret
and correct object detector failures, and Explain Any Concept~\citep{sun2023explain}
combined SAM-style concept regions with concept-based explanation, reflecting a
broader shift from raw pixels to semantically meaningful attribution units.

For insertion and deletion, however, average coalition contribution is not identical to
the best early prefix. A region may have strong average contribution but be redundant
with another early region; conversely, a region may only become decisive in a specific
combination. This gap motivates ordered subset search.

\subsection{Ordered Subset Search}
\label{app:ordered-subset-history}

Ordered subset search directly optimizes the object that insertion and deletion replay.
For an ordering \(\pi\), define \(P_t^\pi=\{\pi_1,\ldots,\pi_t\}\). Instead of first
estimating a static saliency score for each region, a search method chooses regions so
that the prefix response \(G(P_t^\pi)\) recovers quickly for small \(t\). LIMA/Greedy
\citep{chen2024less} follows this principle by repeatedly selecting the remaining
region with the largest marginal gain. Its strength is that each decision is made in
the context of the prefix already chosen; its weakness is the quadratic scan cost
\(n+(n-1)+\cdots+1=O(n^2)\).

VPS~\citep{chen2025vps} extended search-based attribution to object-level foundation
models, where the explained target may be a detection or grounding output rather than
a classification logit. PhaseWin~\citep{gu2026phasewin} made ordered subset search
more scalable by reusing phase-wise structure instead of rescoring every remaining
region after every accepted step. Its extended analysis~\citep{gu2026phasewinExtended}
is especially relevant to this paper: the residual gap to Greedy concentrates on
partition-sensitive tail cases, suggesting that early prefix quality and region
granularity remain central even when the search algorithm is efficient.

This is the immediate predecessor line of our work. It establishes that perturbation
replay and ordered search can outperform map-based attribution when the evaluation
metric is insertion/deletion faithfulness. Our departure is the output object. Many
applications do not need a complete ordering of all \(n\) regions; they need the top
\(k\) decisive regions as a compact evidence mask. \textsc{TRACE} therefore treats the
fixed-cardinality mask itself as the primary object, and uses it as an initialized
prefix only when a downstream benchmark demands a full ordering.

\subsection{Attribution in Multimodal Foundation Models}
\label{app:mllm-history}

The cost and target definition of attribution change sharply in modern multimodal
models. Vision backbones moved from CNNs such as ResNet~\citep{he2016deep} to ViTs
\citep{dosovitskiy2021image} and CLIP-style language-supervised encoders
\citep{radford2021learning}. Multimodal systems such as LLaVA
\citep{liu2023visual,liu2023improved}, BLIP-2~\citep{li2023blip2}, Grounding DINO
\citep{liu2024grounding}, Florence-2~\citep{xiao2024florence}, and Qwen2.5-VL
\citep{bai2025qwen25vl} further change the scoring problem: a candidate mask may have
to be evaluated through a visual encoder, multimodal fusion, region localization,
prompt-conditioned decoding, and autoregressive language generation.

This shift affects both access and semantics. Many high-performing MLLMs expose only a
forward interface, making black-box perturbation more realistic than gradient methods.
Even when gradients are available, the target response is not a single class logit:
captioning, VQA, grounding, and hallucination repair involve token probabilities,
generated strings, object boxes, or corrected answers. Recent MLLM explanation work
reflects this change. MLLMs Know Where to Look~\citep{zhang2025mllms} studied small
visual-detail perception and attention/gradient-based visual intervention. Where do
Large Vision-Language Models Look~\citep{xing2025where} extended heatmap analysis to
open-ended LVLM question answering. TAM~\citep{li2025tam} treated token-level MLLM
activation as the attribution target, while LLaVA-CAM~\citep{zhang2025redundancy} and
IGOS++~\citep{khorram2021igos++} provide strong non-search references in our MLLM
experiments.

EAGLE~\citep{chen2026eagle} is the closest MLLM-side search protocol to our setting:
it adapts greedy region search to autoregressive generation and evaluates both
attribution and repair. POPE~\citep{li2023evaluating} and RePOPE
\citep{neuhaus2025repope} provide the object-hallucination and corrected adversarial
settings in which visual evidence can be tested by decision faithfulness or repair.
In this regime, the central question is no longer whether we can draw a heatmap. The
question is whether we can find a compact visual evidence set that survives behavioral
testing under an acceptable number of expensive multimodal model evaluations.

\subsection{Summary}
\label{app:attribution-history-summary}

The development history can be summarized as a sequence of increasingly behavioral
and combinatorial formulations:
\[
\begin{aligned}
&\text{gradient saliency}
\rightarrow
\text{activation maps}
\rightarrow
\text{perturbation testing}
\rightarrow
\text{coalition contribution}
\\
&\rightarrow
\text{ordered subset search}
\rightarrow
\text{compact evidence search for MLLMs}.
\end{aligned}
\]
Gradient and activation methods solve the efficiency problem but rely on internal
signals. Perturbation and Shapley-style methods test model behavior but often return
independent scores or average contributions. Ordered subset search aligns directly
with insertion/deletion replay but can be expensive when \(n\) is large. \textsc{TRACE}
is positioned at the next step: it keeps the black-box behavioral test, but makes the
compact top-\(k\) evidence mask the primary output, which is the object many
classification, hallucination attribution, and repair applications actually need.

\section{Runtime and Implementation Details}
\label{app:experimental-details}

The main text contains the experimental setup, task definitions, metrics, and MEC
accounting convention. Full numerical tables are reported in
Appendix~\ref{app:additional-experiments}, and supplementary qualitative examples are
included in Section~\ref{app:additional-visualizations}. This appendix records the
additional implementation details needed to reproduce the ablation suite and the MLLM
runtime profiling.

\subsection{Initializer Hyperparameters and Release Convention}
\label{app:init-hparams}

All main experiments use SLICO-64 fine regions. The target response
\(f_y(x)\) is normalized to \([0,1]\), and the initializer optimization score is
the sufficiency proxy \(F(S)=\operatorname{Suff}(S)=f_y(x_S)\) unless explicitly
stated otherwise. Both CoPAIR and TRACE use the same internal fine-region
ordering and release convention: the selected seed candidate is internally
ordered by greedy marginal rescoring, and only the shortest prefix reaching
\(0.8\) of the candidate score is kept as the initialized prefix. The remaining
candidate regions are released to the continuation pool.

For CoPAIR, let \(m=|U|\) be the number of fine regions and \(c\) be the number
of coarse centroid clusters. When no explicit pair-pool size is given, the
candidate pair pool is derived from the pool fraction, minimum size, and reserve:
\[
    q
    =
    \min\left\{
    c,\,
    \max\left(q_{\min},\left\lceil \rho_{\mathrm{pool}}c\right\rceil\right)
    +q_{\mathrm{res}}
    \right\}.
\]
The main CoPAIR settings are listed in Table~\ref{tab:copair-hparams}.

\begin{table}[t]
\centering
\caption{CoPAIR hyperparameters used in the main experiments.}
\label{tab:copair-hparams}
\begin{tabular}{ll}
\toprule
Parameter & Value \\
\midrule
coarse partition & centroid \\
fine superpixels & SLICO-64 \\
adaptive coarse count & \(\operatorname{round}(2.0\sqrt{m})\) \\
coarse count range & \([8,32]\) \\
centroid seed & \(0\) \\
\(k\)-means max iter & \(50\) \\
pair seed pool size & None \\
pair seed pool frac \(\rho_{\mathrm{pool}}\) & \(0.15\) \\
pair seed pool min \(q_{\min}\) & \(10\) \\
pair seed pool reserve \(q_{\mathrm{res}}\) & \(4\) \\
initial score & suff \\
pair min gain \(\delta\) & \(0.05\) \\
min initial sufficiency \(\tau\) & \(0.1\) \\
min initial area frac & \(0.0\) \\
max initial area frac & \(0.2\) \\
max selected coarse regions & \(3\) \\
partial ratio \(\gamma\) & \(0.8\) \\
internal ordering threshold & \(20\) \\
\bottomrule
\end{tabular}
\end{table}

The area cap and the SLICO-64 setting keep the selected fine-region seed below
the internal ordering threshold in the main experiments, so the internal ordering
is exact greedy within the selected seed candidate. If the selected coarse
singleton or pair does not reach the minimum sufficiency threshold, CoPAIR may
absorb a third coarse region by greedy sufficiency improvement, subject to the
maximum selected coarse-region and area-fraction limits.

The main TRACE settings are listed in Table~\ref{tab:trace-hparams}. TRACE uses
Gumbel-top-\(k\) sampling over exactly-\(k\) fine-region masks. The direct-mask
output is the best observed exactly-\(k\) mask. When TRACE is used as an
initializer for full-ordering search, this mask is internally ordered and
partially released using the same \(0.8\) rule as CoPAIR.

\begin{table}[t]
\centering
\caption{TRACE hyperparameters used in the main experiments.}
\label{tab:trace-hparams}
\begin{tabular}{ll}
\toprule
Parameter & Value \\
\midrule
fine superpixels & SLICO-64 \\
seed size \(k\) & classification main: \(8\); POPE/RePOPE: \(10\) \\
rounds \(R\) & \(5\) \\
samples per round \(M\) & \(32\) \\
elite ratio \(\rho_{\mathrm e}\) & \(0.2\) \\
elite count & \(\lceil 0.2\times 32\rceil=7\) \\
update rate \(\eta\) & \(0.7\) \\
optimization score & suff \\
temperature \(T\) & \(1.0\) \\
smoothing \(\lambda_{\mathrm{sm}}\) & \(0.05\) \\
epsilon \(\epsilon\) & \(10^{-4}\) \\
logit clip & \([-8,8]\) \\
random seed & \(0\) \\
suff-only fast scoring & True \\
replay mode for external init & seed only \\
external init key & \texttt{area\_suf\_nec\_selected\_regions} \\
partial ratio \(\gamma\) & \(0.8\) \\
internal ordering threshold & \(20\) \\
\bottomrule
\end{tabular}
\end{table}

Duplicate sampled masks are cached by their sorted region-index tuple. Cached
duplicates reuse the stored score and do not issue additional model forwards.
The MEC values for initialized rows include the uncached initializer calls and
the downstream continuation calls.

\subsection{Ablation Suite Setup}
\label{app:ablation-suite-setup}

The ablations in Appendix~\ref{app:correct-100-ablation} are run on 100 samples drawn
from the first 2,000 correctly classified CLIP ViT-L/14 samples. All images use
SLICO-64 regions~\citep{achanta2012slic}. We report insertion AUC, deletion AUC,
insertion scores at 30\% and 50\% region budgets, the best insertion score along the
curve, and MEC. In this ablation suite, MEC is reported as the raw forward-call count
from the logs, not divided by \(1000\). It includes selection, filtering, and
initialization cost, but excludes the later insertion/deletion replay used only for
evaluation. The ablation jobs were run with 8 shards, and scores are rounded to four
decimals while MEC is rounded to two decimals.

\subsection{Ablated Initializers}
\label{app:ablation-initializers}

The standard \textsc{TRACE} ablation setting is
\(\textsc{TRACE}(k=8,M=32,R=5)+\)Greedy. The \textsc{TRACE} initializer uses
sufficiency scoring, Gumbel-top-\(k\) sampling with temperature \(1.0\), elite ratio
\(0.2\), EMA update rate \(\eta=0.7\), smoothing \(0.05\), logit clipping \(8.0\), and
random seed \(0\). After the final round, the best sampled set is used as the seed set
for the downstream Greedy ordering; selected regions are placed first and the remaining
regions are ranked by the learned logits before continuation.

\textsc{CoPAIR} uses centroid coarsening. When the target coarse-region count is set
to \(0\), it is resolved adaptively from the fine-region count with minimum 8, maximum
32, and scale \(2.0\). Coarse seed candidates are selected by sufficiency. The seed
pool uses a 0.15 top-group fraction with minimum size 10. The default thresholds are
minimum initial sufficiency \(0.1\), pair minimum gain \(0.05\), maximum initial
fraction \(0.2\), and partial-release ratio \(0.8\). After the initial coarse seed is
chosen, the selected fine regions are partially ordered and released according to the
same initialization interface used in Section~\ref{subsec:copair}.

\subsection{Memory-Saving Settings for Runtime Profiling}
\label{app:memory-saving-settings}

For the timing and memory results in Table~\ref{tab:direct-repair-main}, LLaVA-CAM and
IGOS++ were deployed on an RTX 5090 using memory-saving settings to make the comparison
easy to reproduce on widely available hardware. The original, uncompressed peak-memory
measurements reported outside the parentheses were run on an RTX PRO 6000. The
parenthesized engineered measurements were obtained on the 5090 with the following
settings.
\begin{itemize}[leftmargin=1.35em,itemsep=1pt,topsep=2pt]
\raggedright
\item \textbf{Profiling isolation.} We used \texttt{score\_batch\_size=1}, one profiling process per method/sample stream, and \texttt{PYTORCH\_CUDA\_ALLOC\_CONF=}\texttt{expandable\_segments:True}. Before each measured method call, the profiler ran Python garbage collection, \texttt{torch.cuda.empty\_cache()}, CUDA synchronization, and peak-memory reset, so the reported peak corresponds to the current method execution under the already loaded model.
\item \textbf{Common forward settings.} All scoring, generation, and gradient-attribution forward passes used \texttt{use\_cache=False}, so KV-cache memory was not retained during attribution. The segmentation and search configuration was fixed to SLICO-64 with \texttt{target\_length=64}.
\item \textbf{Qwen2.5-VL-3B.} We used \texttt{bfloat16}, FlashAttention-2 when available, and image-size caps of max image side 448, min pixels 3136, and max pixels 401408.
\item \textbf{LLaVA-v1.5-7B.} We used \texttt{float16}, FlashAttention-2 when available, frozen model parameters via \texttt{requires\_grad\_(False)}, gradient checkpointing, \texttt{model.config.use\_cache=False}, and image-size caps of max image side 336, min pixels 3136, and max pixels 200704. These LLaVA-specific settings are the memory-saving changes that allowed LLaVA-CAM and IGOS++ to run on the 5090.
\end{itemize}

\section{Additional Results and Ablations}
\label{app:additional-experiments}

This appendix provides the detailed tables that support the compact presentation in
Section~\ref{sec:experiments}. We first report a focused ablation suite on CLIP
ViT-L/14 Correct samples with SLICO-64 regions. This subset is intended for
method-design questions, not for claiming the final leaderboard. We then report
additional metrics for the remaining classification backbones, full MLLM attribution
metrics, and the full repair protocol results. Implementation and cost-accounting
details are reported in Appendix~\ref{app:experimental-details}; qualitative examples
are included in Section~\ref{app:additional-visualizations}.

\subsection{CLIP ViT-L/14 Correct-100 Ablation Suite}
\label{app:correct-100-ablation}

All ablations in this subsection use the same 100-sample subset drawn from correctly
classified CLIP ViT-L/14 samples. The purpose is to isolate method-design choices under
a small, reproducible evaluation budget before running full-scale tables. Unless
otherwise stated, metrics follow the ordered-attribution setup in
Section~\ref{subsec:metrics}: InsAUC is the primary faithfulness metric for
Correct, DelAUC and High are secondary, and MEC is the raw selection/filter/initializer
call count excluding curve replay.

\subsubsection{Initializer Comparison}
\label{app:ablation-initializer-comparison}

Table~\ref{tab:app-ablate-initializers} compares raw Greedy, random fixed-size seeds,
relaxed ASN seeds, \textsc{CoPAIR}, and the standard
\(\textsc{TRACE}(k=8,M=32,R=5)+\)Greedy configuration. \textsc{TRACE} gives the best
InsAUC, @30, @50, and High values. \textsc{CoPAIR} is the strongest non-\textsc{TRACE}
initializer and substantially improves over Greedy. Random and ASN seeds reduce MEC
but do not improve insertion faithfulness, indicating that the gain is not explained
by simply shortening the continuation problem.

The ASN rows are included as a targeted negative control for traditional relaxed-mask
initialization. ASN optimizes continuous region variables under a combined sufficiency
and necessity objective,
\[
    \operatorname{SN}
    =
    \alpha\,\operatorname{Suff}
    +
    (1-\alpha)\,\operatorname{Necc},
\]
then hardens the relaxed mask into a seed for Greedy. We test a fixed top-10 rule, an
SN-knee rule over sizes 5--16, and an SN-plateau rule over the same range. All three
use \texttt{dark\_defocus} perturbations, \texttt{topk\_suff} objective,
\(\gamma_s=\gamma_n=0.8\), score margin, 25 optimization steps, area budget \(0.6\),
\(\mu=0.01\), \(\nu=0.05\), and \(\lambda_{\mathrm{area}}=0.0\). Their weaker InsAUC
shows that a conventional relaxed mask, even when converted into a Greedy seed, does
not substitute for discrete fixed-\(k\) TRACE search.

\begin{table}[ht]
\centering
\caption{Initializer comparison on the CLIP ViT-L/14 Correct-100 ablation subset. MEC is reported as raw forward-call count and excludes insertion/deletion replay.}
\label{tab:app-ablate-initializers}
\scriptsize
\setlength{\tabcolsep}{3.5pt}
\resizebox{\linewidth}{!}{
\begin{tabular}{lcccccc}
\toprule
Method & InsAUC & DelAUC & @30 & @50 & High & MEC \\
\midrule
Greedy & 0.8386 & 0.1081 & 0.9010 & 0.9532 & 0.9743 & 4111.64 \\
Random-\(k\) seed 0 + Greedy & 0.8248 & 0.1285 & 0.8885 & 0.9578 & 0.9771 & 3733.98 \\
Random-\(k\) seed 1 + Greedy & 0.8268 & 0.1432 & 0.9042 & 0.9583 & 0.9757 & 3690.76 \\
Random-\(k\) seed 2 + Greedy & 0.8261 & 0.1321 & 0.8974 & 0.9588 & 0.9801 & 3741.90 \\
ASN fixed-\(k10\) + Greedy & 0.8210 & 0.1311 & 0.8435 & 0.9581 & 0.9771 & 3530.52 \\
ASN SN-knee 5--16 + Greedy & 0.8124 & 0.1182 & 0.8073 & 0.9531 & 0.9771 & 3450.74 \\
ASN SN-plateau 5--16 + Greedy & 0.8134 & 0.1194 & 0.8069 & 0.9511 & 0.9765 & 3452.28 \\
\textsc{CoPAIR} + Greedy & 0.8652 & 0.1245 & 0.9368 & 0.9688 & 0.9772 & 3653.48 \\
\textsc{TRACE} + Greedy & \bestcell{0.8718} & 0.1197 & \bestcell{0.9466} & \bestcell{0.9698} & \bestcell{0.9802} & 3733.16 \\
\bottomrule
\end{tabular}}
\end{table}

\subsubsection{TRACE Sampling Budget}
\label{app:ablation-trace-budget}

Table~\ref{tab:app-ablate-trace-budget} varies the number of sampled masks per round
and the number of cross-entropy rounds. Increasing samples from 8 to 32 gives a clear
InsAUC gain. Increasing to 64 gives only a small additional gain while increasing MEC.
For rounds, moving from 1 to 5 improves InsAUC, while 8 rounds yields nearly the same
quality as larger sampling budgets at additional cost. The \(M=32,R=5\) configuration
is therefore used as the balanced default.

\begin{table}[ht]
\centering
\caption{\textsc{TRACE} sample/round ablation on the CLIP ViT-L/14 Correct-100 subset.}
\label{tab:app-ablate-trace-budget}
\scriptsize
\setlength{\tabcolsep}{4pt}
\resizebox{0.82\linewidth}{!}{
\begin{tabular}{lcccccc}
\toprule
Method & InsAUC & DelAUC & @30 & @50 & High & MEC \\
\midrule
\textsc{TRACE} \(M=8,R=5\) + Greedy & 0.8408 & 0.1380 & 0.9283 & 0.9621 & 0.9748 & 3549.95 \\
\textsc{TRACE} \(M=16,R=5\) + Greedy & 0.8539 & 0.1302 & 0.9246 & 0.9638 & 0.9782 & 3624.57 \\
\textsc{TRACE} \(M=32,R=5\) + Greedy & 0.8718 & 0.1197 & 0.9466 & 0.9698 & 0.9802 & 3733.16 \\
\textsc{TRACE} \(M=64,R=5\) + Greedy & 0.8734 & 0.1220 & 0.9429 & 0.9612 & 0.9765 & 3921.17 \\
\textsc{TRACE} \(M=32,R=1\) + Greedy & 0.8456 & 0.1304 & 0.9342 & 0.9680 & 0.9776 & 3589.16 \\
\textsc{TRACE} \(M=32,R=3\) + Greedy & 0.8632 & 0.1261 & 0.9362 & 0.9663 & 0.9771 & 3662.00 \\
\textsc{TRACE} \(M=32,R=8\) + Greedy & 0.8733 & 0.1167 & 0.9424 & 0.9665 & 0.9780 & 3773.28 \\
\bottomrule
\end{tabular}}
\end{table}

\subsubsection{TRACE Mask Size}
\label{app:ablation-trace-k}

Table~\ref{tab:app-ablate-trace-k} varies the fixed mask size \(k\). The default
\(k=8\) gives the best InsAUC, @50, and High values. Smaller masks are close but
slightly weaker; larger masks reduce MEC because more regions are initialized before
continuation, but they also reduce InsAUC. This supports treating \(k\) as an output
contract rather than a free cost-reduction knob.

\begin{table}[ht]
\centering
\caption{Fixed-cardinality mask-size ablation on the CLIP ViT-L/14 Correct-100 subset.}
\label{tab:app-ablate-trace-k}
\scriptsize
\setlength{\tabcolsep}{4pt}
\resizebox{0.76\linewidth}{!}{
\begin{tabular}{lcccccc}
\toprule
Method & InsAUC & DelAUC & @30 & @50 & High & MEC \\
\midrule
\textsc{TRACE} \(k=4\) + Greedy & 0.8683 & 0.1243 & 0.9245 & 0.9651 & 0.9774 & 3844.96 \\
\textsc{TRACE} \(k=6\) + Greedy & 0.8700 & 0.1236 & 0.9321 & 0.9609 & 0.9750 & 3789.99 \\
\textsc{TRACE} \(k=8\) + Greedy & \bestcell{0.8718} & 0.1197 & \bestcell{0.9466} & \bestcell{0.9698} & \bestcell{0.9802} & 3733.16 \\
\textsc{TRACE} \(k=10\) + Greedy & 0.8655 & 0.1160 & 0.9352 & 0.9630 & 0.9767 & 3709.14 \\
\textsc{TRACE} \(k=12\) + Greedy & 0.8599 & 0.1139 & 0.9282 & 0.9607 & 0.9766 & 3682.94 \\
\textsc{TRACE} \(k=16\) + Greedy & 0.8612 & 0.1113 & 0.9421 & 0.9682 & 0.9769 & 3641.58 \\
\bottomrule
\end{tabular}}
\end{table}

\subsubsection{Mask-Size Overlap}
\label{app:ablation-trace-overlap}

Table~\ref{tab:app-ablate-trace-overlap} compares the selected regions across
different \(k\) values. For example, \texttt{top4\_of\_k6\_vs\_k4} measures the overlap
between the top four regions from a \(k=6\) run and the selected \(k=4\) mask. The
overlaps are moderate rather than near-identical, so changing \(k\) does not merely
append extra regions to a stable core set; it can change the seed composition.

\begin{table}[ht]
\centering
\caption{Overlap among \textsc{TRACE} masks selected with different fixed cardinalities.}
\label{tab:app-ablate-trace-overlap}
\scriptsize
\setlength{\tabcolsep}{8pt}
\begin{tabular}{lcc}
\toprule
Comparison & Overlap & Jaccard \\
\midrule
\texttt{top4\_of\_k6\_vs\_k4} & 0.4325 & 0.2991 \\
\texttt{top4\_of\_k8\_vs\_k4} & 0.4275 & 0.2987 \\
\texttt{top4\_of\_k10\_vs\_k4} & 0.3600 & 0.2439 \\
\texttt{top4\_of\_k12\_vs\_k4} & 0.3125 & 0.2078 \\
\texttt{top4\_of\_k16\_vs\_k4} & 0.2800 & 0.1921 \\
\texttt{top6\_of\_k8\_vs\_k6} & 0.4650 & 0.3241 \\
\texttt{top6\_of\_k10\_vs\_k6} & 0.4183 & 0.2773 \\
\texttt{top6\_of\_k12\_vs\_k6} & 0.3933 & 0.2565 \\
\texttt{top6\_of\_k16\_vs\_k6} & 0.3400 & 0.2165 \\
\texttt{top8\_of\_k10\_vs\_k8} & 0.4613 & 0.3119 \\
\texttt{top8\_of\_k12\_vs\_k8} & 0.4425 & 0.2984 \\
\texttt{top8\_of\_k16\_vs\_k8} & 0.3650 & 0.2317 \\
\texttt{top10\_of\_k12\_vs\_k10} & 0.4800 & 0.3255 \\
\texttt{top10\_of\_k16\_vs\_k10} & 0.4230 & 0.2805 \\
\texttt{top12\_of\_k16\_vs\_k12} & 0.4708 & 0.3198 \\
\bottomrule
\end{tabular}
\end{table}

\subsection{CLIP RN101 Fixed-\texorpdfstring{\(k\)}{k} Extension}
\label{app:clip-rn101-fixed-k}

The CLIP RN101 block in Table~\ref{tab:classification-search-main} is the main case
where \textsc{CoPAIR}+Greedy is slightly stronger than the default
\(\textsc{TRACE}(k=8)+\)Greedy configuration. We attribute this to the convolutional
ResNet visual encoder: compared with ViT patch tokens, RN101 features more readily
align broad spatial regions with the CLIP semantic space. This makes insertion curves
partly redundant and weakens the insertion/deletion agreement of a small fixed-size
seed. To test whether this is a failure of \textsc{TRACE} or a mismatch in the output
cardinality, we reran the CLIP RN101 comparison with larger \textsc{TRACE} masks. All
\textsc{TRACE} rows use the same inner partial-greedy completion before the backend
Greedy continuation, and we abbreviate this pipeline as
\(\textsc{TRACE}(k)+\)Greedy.

Table~\ref{tab:app-clip-rn101-k-extension} shows that increasing \(k\) consistently
improves deletion behavior and reduces MEC relative to \(k=8\). On Cause and Repair,
\(k=16\) already exceeds \textsc{CoPAIR}+Greedy on the primary metric for the split
(InsAUC for Cause and High for Repair). Correct remains harder because its positive
evidence often covers larger semantic regions: \textsc{CoPAIR} still has the best
InsAUC, while larger \textsc{TRACE} masks improve DelAUC, High, and cost. The
Correct-only \(k=12,16,20\) sweep also matches the trend in
Table~\ref{tab:app-ablate-trace-k}: larger fixed masks shorten the remaining Greedy
continuation and lower MEC, without making \textsc{TRACE} collapse as the number of
selected regions grows.

\begin{table}[ht]
\centering
\caption{CLIP RN101 larger-mask extension. Lower DelAUC is better. Gold cells mark the best non-cost metric within each split.}
\label{tab:app-clip-rn101-k-extension}
\scriptsize
\setlength{\tabcolsep}{2.5pt}
\resizebox{\linewidth}{!}{
\begin{tabular}{llccccccc}
\toprule
Split & Method & \(n\) & InsAUC & DelAUC & @30 & @50 & High & MEC \\
\midrule
Correct & \textsc{CoPAIR}+Greedy & 500 & \bestcell{0.7017} & 0.0751 & \bestcell{0.7423} & \bestcell{0.8503} & 0.9105 & 3406.28 \\
Correct & \(\textsc{TRACE}(k=8)\)+Greedy & 500 & 0.6797 & 0.0720 & 0.6805 & 0.8231 & 0.9051 & 3564.67 \\
Correct & \(\textsc{TRACE}(k=12)\)+Greedy & 500 & 0.6841 & 0.0691 & 0.7016 & 0.8357 & 0.9084 & 3402.13 \\
Correct & \(\textsc{TRACE}(k=16)\)+Greedy & 500 & 0.6825 & 0.0640 & 0.6960 & 0.8449 & 0.9093 & 3281.83 \\
Correct & \(\textsc{TRACE}(k=20)\)+Greedy & 500 & 0.6813 & \bestcell{0.0580} & 0.6860 & 0.8499 & \bestcell{0.9149} & 3220.54 \\
\midrule
Cause & \textsc{CoPAIR}+Greedy & 100 & 0.5722 & 0.0405 & 0.6030 & 0.7283 & 0.8037 & 3411.68 \\
Cause & \(\textsc{TRACE}(k=8)\)+Greedy & 100 & 0.5597 & 0.0408 & 0.6016 & 0.7006 & 0.8027 & 3594.89 \\
Cause & \(\textsc{TRACE}(k=16)\)+Greedy & 100 & \bestcell{0.5741} & \bestcell{0.0291} & \bestcell{0.6074} & \bestcell{0.7329} & \bestcell{0.8176} & 3374.04 \\
\midrule
Repair & \textsc{CoPAIR}+Greedy & 100 & 0.3435 & 0.0129 & 0.4097 & 0.5053 & 0.5653 & 3428.30 \\
Repair & \(\textsc{TRACE}(k=8)\)+Greedy & 100 & 0.3494 & 0.0116 & 0.4141 & \bestcell{0.5113} & 0.5679 & 3653.70 \\
Repair & \(\textsc{TRACE}(k=16)\)+Greedy & 100 & \bestcell{0.3502} & \bestcell{0.0112} & \bestcell{0.4253} & 0.5067 & \bestcell{0.5766} & 3393.96 \\
\bottomrule
\end{tabular}}
\end{table}

\subsection{Full Classification Tables for Additional Backbones}
\label{app:classification-full}

Tables~\ref{tab:app-clip-rn101-full} and~\ref{tab:app-resnet101-full} report the full classification metrics for CLIP ResNet-101 and supervised ResNet-101. Correct and Cause are interpreted primarily through InsAUC, while Repair is interpreted primarily through Highest.

\begin{table}[ht]
\centering
\caption{Full classification attribution results on CLIP ResNet-101 using ImageNet-derived splits.}
\label{tab:app-clip-rn101-full}
\scriptsize
\setlength{\tabcolsep}{2pt}
\resizebox{\linewidth}{!}{
\begin{tabular}{lccccccccc}
\toprule
Method & \multicolumn{3}{c}{Correct} & \multicolumn{3}{c}{Cause} & \multicolumn{3}{c}{Repair} \\
\cmidrule(lr){2-4}\cmidrule(lr){5-7}\cmidrule(lr){8-10}
 & InsAUC & DelAUC & Highest & InsAUC & DelAUC & Highest & InsAUC & DelAUC & Highest \\
\midrule
RISE & 0.4502 & 0.1107 & 0.7963 & 0.2492 & 0.0514 & 0.5487 & 0.0967 & 0.0148 & 0.2470 \\
HSIC & 0.4335 & 0.1158 & 0.7796 & 0.2296 & 0.0590 & 0.5232 & 0.0789 & 0.0182 & 0.2114 \\
Gradient & 0.3392 & 0.2097 & 0.7727 & 0.1812 & 0.1069 & 0.5147 & 0.0575 & 0.0331 & 0.1894 \\
IG2 & 0.3431 & 0.2190 & 0.7748 & 0.1862 & 0.1099 & 0.5157 & 0.0583 & 0.0361 & 0.1902 \\
IGOS++ & 0.2592 & 0.1932 & 0.7664 & 0.1347 & 0.0970 & 0.5119 & 0.0477 & 0.0309 & 0.1868 \\
\midrule
Greedy & 0.6677 & 0.0462 & 0.9134 & 0.5163 & 0.0223 & 0.7780 & 0.3195 & 0.0092 & 0.5478 \\
PhaseWin & 0.5691 & 0.0500 & 0.8886 & 0.4087 & 0.0230 & 0.7366 & 0.2265 & 0.0090 & 0.4657 \\
\textsc{CoPAIR}+PhaseWin & 0.6688 & 0.0783 & 0.8975 & 0.5019 & 0.0378 & 0.7505 & 0.2983 & 0.0122 & 0.5170 \\
\textsc{TRACE}+PhaseWin & 0.6376 & 0.0750 & 0.8928 & 0.4753 & 0.0377 & 0.7437 & 0.2908 & 0.0117 & 0.5073 \\
\textsc{CoPAIR}+Greedy & \bestcell{0.7038} & 0.0746 & 0.9134 & \bestcell{0.5454} & 0.0360 & \bestcell{0.7824} & \bestcell{0.3348} & 0.0119 & \bestcell{0.5595} \\
\textsc{TRACE}+Greedy & 0.6852 & 0.0718 & 0.9102 & 0.5260 & 0.0373 & 0.7708 & 0.3344 & 0.0116 & 0.5563 \\
\bottomrule
\end{tabular}}
\end{table}

\begin{table}[ht]
\centering
\caption{Full classification attribution results on supervised ResNet-101 using ImageNet-derived splits.}
\label{tab:app-resnet101-full}
\scriptsize
\setlength{\tabcolsep}{2pt}
\resizebox{\linewidth}{!}{
\begin{tabular}{lccccccccc}
\toprule
Method & \multicolumn{3}{c}{Correct} & \multicolumn{3}{c}{Cause} & \multicolumn{3}{c}{Repair} \\
\cmidrule(lr){2-4}\cmidrule(lr){5-7}\cmidrule(lr){8-10}
 & InsAUC & DelAUC & Highest & InsAUC & DelAUC & Highest & InsAUC & DelAUC & Highest \\
\midrule
RISE & 0.6218 & 0.2477 & 0.8359 & 0.3789 & 0.1139 & 0.6430 & 0.1615 & 0.0339 & 0.3641 \\
HSIC & 0.6248 & 0.2271 & 0.8324 & 0.3645 & 0.1152 & 0.6202 & 0.1094 & 0.0444 & 0.2908 \\
Gradient & 0.4658 & 0.3746 & 0.8069 & 0.2774 & 0.1857 & 0.5830 & 0.0937 & 0.0526 & 0.2638 \\
IGOS++ & 0.4841 & 0.3564 & 0.8189 & 0.2761 & 0.1754 & 0.6115 & 0.1036 & 0.0513 & 0.2832 \\
\midrule
Greedy & 0.8407 & 0.1301 & 0.9605 & 0.7390 & 0.0479 & \bestcell{0.9155} & 0.5482 & 0.0173 & 0.7758 \\
PhaseWin & 0.8043 & 0.1483 & 0.9462 & 0.6802 & 0.0536 & 0.8883 & 0.4670 & 0.0185 & 0.7067 \\
\textsc{CoPAIR}+PhaseWin & 0.8145 & 0.1630 & 0.9469 & 0.6971 & 0.0659 & 0.8901 & 0.5017 & 0.0214 & 0.7479 \\
\textsc{TRACE}+PhaseWin & 0.8214 & 0.1581 & 0.9487 & 0.7035 & 0.0648 & 0.8916 & 0.5120 & 0.0214 & 0.7511 \\
\textsc{CoPAIR}+Greedy & 0.8394 & 0.1451 & 0.9602 & 0.7346 & 0.0606 & 0.9142 & 0.5377 & 0.0206 & 0.7778 \\
\textsc{TRACE}+Greedy & \bestcell{0.8478} & 0.1420 & \bestcell{0.9616} & \bestcell{0.7414} & 0.0597 & 0.9139 & \bestcell{0.5490} & 0.0201 & \bestcell{0.7794} \\
\bottomrule
\end{tabular}}
\end{table}

\subsection{Additional CLIP RN101 Visualizations}
\label{app:additional-visualizations}
\label{app:additional-classification-visualizations}

\begin{figure}[ht]
\centering
\includegraphics[width=\linewidth]{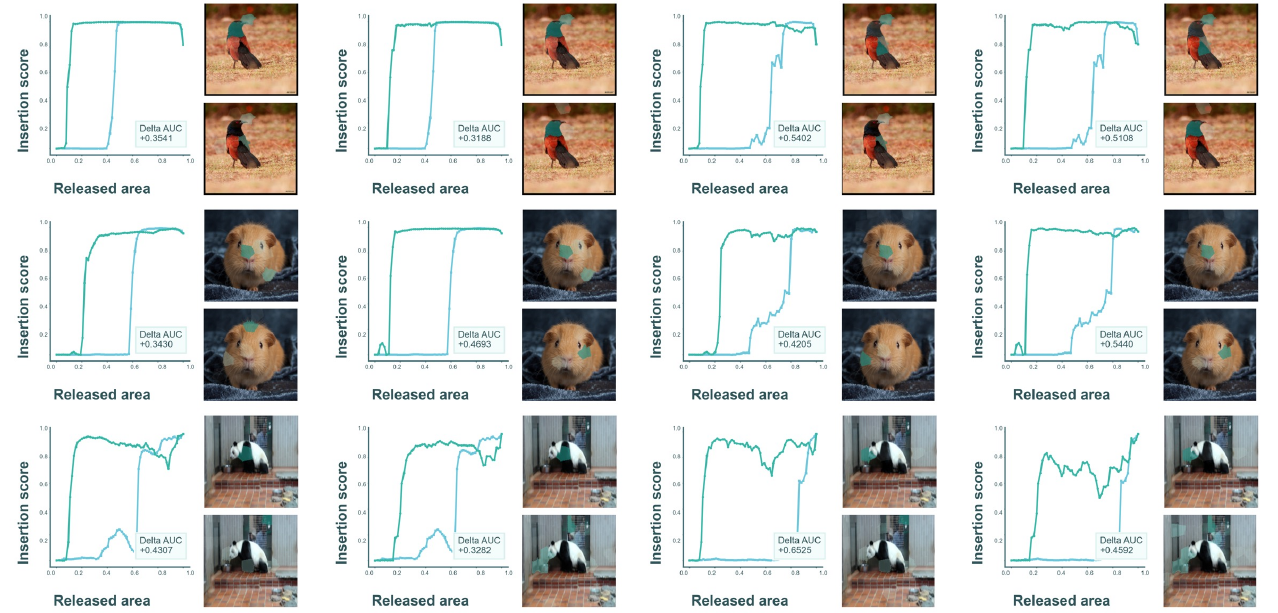}
\caption{Qualitative CLIP RN101 Correct-split examples complementing
Table~\ref{tab:classification-search-main}, Table~\ref{tab:app-clip-rn101-full}, and
Table~\ref{tab:app-clip-rn101-k-extension}. The visualizations show representative
insertion curves and evidence masks for the additional CLIP backbone, mirroring the
CLIP ViT-L/14 qualitative analysis in Figure~\ref{fig:clip-vit-correct-vis}.}
\label{fig:app-clip-rn101-correct-vis}
\end{figure}

Figure~\ref{fig:app-clip-rn101-correct-vis} provides representative CLIP RN101
Correct-split cases corresponding to the CLIP RN101 block in
Table~\ref{tab:classification-search-main}, the full metrics in
Table~\ref{tab:app-clip-rn101-full}, and the larger-mask check in
Table~\ref{tab:app-clip-rn101-k-extension}. These examples mirror the CLIP ViT-L/14
qualitative analysis in Figure~\ref{fig:clip-vit-correct-vis}.

\subsection{Full MLLM Attribution Metrics}
\label{app:mllm-full-metrics}

Tables~\ref{tab:app-qwen-mllm-full} and~\ref{tab:app-llava-mllm-full} report the full POPE and RePOPE attribution metrics for the two MLLMs. Forward calls are reported only for forward-only methods; backward and optimization-based methods are marked with ``--'' in the cost column.

\begin{table}[ht]
\centering
\caption{Full Qwen2.5-VL-3B attribution metrics on POPE and RePOPE.}
\label{tab:app-qwen-mllm-full}
\scriptsize
\setlength{\tabcolsep}{3pt}
\resizebox{\linewidth}{!}{
\begin{tabular}{llcccccc}
\toprule
Split & Method & AUC & @30 & @50 & Highest & DelAUC & Forward calls \\
\midrule
POPE & LLaVA-CAM & 0.7652 & 0.7215 & 0.7869 & 0.9506 & 0.7736 & -- \\
POPE & IGOS++ & 0.7754 & 0.7267 & 0.8163 & 0.9504 & 0.7542 & -- \\
POPE & Greedy & 0.9305 & 0.9346 & 0.9462 & 0.9750 & 0.3707 & 4148 \\
POPE & PhaseWin & 0.9253 & 0.9283 & 0.9374 & 0.9735 & 0.4159 & 2239 \\
POPE & \textsc{CoPAIR}+PhaseWin & 0.9308 & 0.9371 & 0.9434 & 0.9742 & 0.4188 & 2206 \\
POPE & \textsc{CoPAIR}+Greedy & 0.9347 & 0.9431 & 0.9485 & 0.9763 & 0.3693 & 4094 \\
POPE & \textsc{TRACE}+PhaseWin & 0.9299 & 0.9348 & 0.9393 & 0.9748 & 0.4213 & 2098 \\
POPE & \textsc{TRACE}+Greedy & 0.9353 & 0.9423 & 0.9476 & 0.9758 & 0.3674 & 3984 \\
POPE & \textsc{TRACE} Seed/Direct & 0.8908 & 0.9226 & 0.8828 & 0.9701 & 0.6849 & 164 \\
\midrule
RePOPE & LLaVA-CAM & 0.3745 & 0.3917 & 0.3683 & 0.6873 & 0.3919 & -- \\
RePOPE & IGOS++ & 0.3721 & 0.4052 & 0.3791 & 0.6916 & 0.3874 & -- \\
RePOPE & Greedy & 0.8055 & 0.8751 & 0.8593 & 0.9245 & 0.1284 & 4092 \\
RePOPE & PhaseWin & 0.7783 & 0.8342 & 0.8184 & 0.9122 & 0.1424 & 2061 \\
RePOPE & \textsc{CoPAIR}+PhaseWin & 0.7820 & 0.8459 & 0.8268 & 0.9148 & 0.1458 & 2054 \\
RePOPE & \textsc{CoPAIR}+Greedy & 0.8028 & 0.8768 & 0.8560 & 0.9229 & 0.1302 & 4009 \\
RePOPE & \textsc{TRACE}+PhaseWin & 0.7810 & 0.8385 & 0.8216 & 0.9179 & 0.1497 & 1931 \\
RePOPE & \textsc{TRACE}+Greedy & 0.8045 & 0.8830 & 0.8561 & 0.9273 & 0.1322 & 3878 \\
RePOPE & \textsc{TRACE} Seed/Direct & 0.5882 & 0.7590 & 0.5763 & 0.8822 & 0.3018 & 164 \\
RePOPE & \textsc{CoPAIR} Direct & 0.6421 & 0.6421 & 0.6421 & 0.6421 & 0.2307 & 142 \\
RePOPE & \textsc{TRACE} Direct & 0.7633 & 0.8639 & 0.8639 & 0.8718 & 0.2875 & 165 \\
\bottomrule
\end{tabular}}
\end{table}

\begin{table}[ht]
\centering
\caption{Full LLaVA-v1.5-7B attribution metrics on POPE and RePOPE.}
\label{tab:app-llava-mllm-full}
\scriptsize
\setlength{\tabcolsep}{3pt}
\resizebox{\linewidth}{!}{
\begin{tabular}{llcccccc}
\toprule
Split & Method & AUC & @30 & @50 & Highest & DelAUC & Forward calls \\
\midrule
POPE & LLaVA-CAM & 0.7563 & 0.7043 & 0.7861 & 0.9270 & 0.7735 & -- \\
POPE & IGOS++ & 0.7879 & 0.7559 & 0.8133 & 0.9277 & 0.7594 & -- \\
POPE & Greedy & 0.9152 & 0.9208 & 0.9216 & 0.9484 & 0.4841 & 4255 \\
POPE & PhaseWin & 0.9090 & 0.9119 & 0.9161 & 0.9466 & 0.5254 & 1543 \\
POPE & \textsc{CoPAIR}+PhaseWin & 0.9104 & 0.9182 & 0.9174 & 0.9466 & 0.5226 & 1542 \\
POPE & \textsc{CoPAIR}+Greedy & 0.9161 & 0.9274 & 0.9243 & 0.9495 & 0.4738 & 4218 \\
POPE & \textsc{TRACE}+PhaseWin & 0.9102 & 0.9192 & 0.9134 & 0.9481 & 0.5216 & 1434 \\
POPE & \textsc{TRACE}+Greedy & 0.9165 & 0.9272 & 0.9233 & 0.9499 & 0.4693 & 4102 \\
POPE & \textsc{TRACE} Seed/Direct & 0.8859 & 0.9119 & 0.8852 & 0.9450 & 0.7045 & 164 \\
\midrule
RePOPE & LLaVA-CAM & 0.3697 & 0.3951 & 0.3726 & 0.6737 & 0.3628 & -- \\
RePOPE & IGOS++ & 0.3592 & 0.3886 & 0.3524 & 0.6815 & 0.3784 & -- \\
RePOPE & Greedy & 0.7683 & 0.8410 & 0.8132 & 0.8982 & 0.1603 & 4241 \\
RePOPE & PhaseWin & 0.7487 & 0.8095 & 0.7760 & 0.8893 & 0.1719 & 1708 \\
RePOPE & \textsc{CoPAIR}+PhaseWin & 0.7553 & 0.8337 & 0.7846 & 0.8925 & 0.1719 & 1642 \\
RePOPE & \textsc{CoPAIR}+Greedy & 0.7779 & 0.8570 & 0.8227 & 0.9042 & 0.1606 & 4162 \\
RePOPE & \textsc{TRACE}+PhaseWin & 0.7620 & 0.8417 & 0.7982 & 0.8959 & 0.1721 & 1458 \\
RePOPE & \textsc{TRACE}+Greedy & 0.7859 & 0.8742 & 0.8391 & 0.9082 & 0.1589 & 3994 \\
RePOPE & \textsc{TRACE} Seed/Direct & 0.5703 & 0.7170 & 0.5737 & 0.8580 & 0.2844 & 164 \\
\bottomrule
\end{tabular}}
\end{table}

\subsection{Full EAGLE Repair Protocol Results}
\label{app:eagle-repair-results}

Table~\ref{tab:app-eagle-repair} reports the full keep-prefix repair results under the protocol from \emph{Where MLLMs Attend and What They Rely On: Explaining Autoregressive Token Generation}. AMCR and CSR@10 are defined only for ordered trajectories. Direct \textsc{CoPAIR}/\textsc{TRACE} masks are therefore reported only by repair rate, failed sample count, and MEC where available.

\begin{table}[ht]
\centering
\caption{Full repair protocol results on RePOPE following \emph{Where MLLMs Attend and What They Rely On: Explaining Autoregressive Token Generation}. Direct mask rows have no ordering, so AMCR and CSR@10 are not applicable.}
\label{tab:app-eagle-repair}
\scriptsize
\setlength{\tabcolsep}{3pt}
\resizebox{\linewidth}{!}{
\begin{tabular}{lllccccc}
\toprule
Model & Method & Protocol & AMCR & CSR@10 & Repair Rate & Failed & MEC \\
\midrule
Qwen2.5-VL-3B & LLaVA-CAM & full curve & 0.4638 & 0.3778 & 0.6963 & 82 & -- \\
Qwen2.5-VL-3B & IGOS++ & full curve & 0.4754 & 0.3556 & 0.7259 & 74 & -- \\
Qwen2.5-VL-3B & Greedy & full curve & 0.1092 & 0.7111 & 0.9926 & 2 & 4092 \\
Qwen2.5-VL-3B & PhaseWin & full curve & \bestcell{0.0981} & 0.7074 & \bestcell{0.9963} & \bestcell{1} & 2061 \\
Qwen2.5-VL-3B & \textsc{CoPAIR}+PhaseWin & full curve & 0.1100 & 0.7111 & 0.9926 & 2 & 2054 \\
Qwen2.5-VL-3B & \textsc{CoPAIR}+Greedy & full curve & 0.1051 & \bestcell{0.7630} & 0.9926 & 2 & 4009 \\
Qwen2.5-VL-3B & \textsc{TRACE}+PhaseWin & full curve & 0.1100 & 0.7148 & 0.9926 & 2 & 1931 \\
Qwen2.5-VL-3B & \textsc{TRACE}+Greedy & full curve & 0.1073 & 0.7148 & 0.9889 & 3 & 3878 \\
Qwen2.5-VL-3B & \textsc{CoPAIR} Direct & single point & -- & -- & 0.8444 & 42 & 142 \\
Qwen2.5-VL-3B & \textsc{TRACE} Direct & single point & -- & -- & 0.9444 & 15 & 165 \\
\midrule
LLaVA-v1.5-7B & LLaVA-CAM & full curve & 0.6032 & 0.1033 & 0.6833 & 95 & -- \\
LLaVA-v1.5-7B & IGOS++ & full curve & 0.6365 & 0.0900 & 0.7033 & 89 & -- \\
LLaVA-v1.5-7B & Greedy & full curve & 0.1240 & 0.6700 & 0.9933 & 2 & 4241 \\
LLaVA-v1.5-7B & PhaseWin & full curve & 0.1168 & 0.6967 & 0.9900 & 3 & 1708 \\
LLaVA-v1.5-7B & \textsc{CoPAIR}+PhaseWin & full curve & 0.1292 & 0.6900 & 0.9933 & 2 & 1642 \\
LLaVA-v1.5-7B & \textsc{CoPAIR}+Greedy & full curve & 0.1131 & 0.7000 & \bestcell{1.0000} & \bestcell{0} & 4162 \\
LLaVA-v1.5-7B & \textsc{TRACE}+PhaseWin & full curve & 0.1123 & 0.6833 & \bestcell{1.0000} & \bestcell{0} & 1458 \\
LLaVA-v1.5-7B & \textsc{TRACE}+Greedy & full curve & \bestcell{0.1093} & \bestcell{0.7167} & \bestcell{1.0000} & \bestcell{0} & 3994 \\
LLaVA-v1.5-7B & \textsc{CoPAIR} Direct & single point & -- & -- & 0.8267 & 52 & -- \\
LLaVA-v1.5-7B & \textsc{TRACE} Direct & single point & -- & -- & 0.9600 & 12 & 164 \\
\bottomrule
\end{tabular}}
\end{table}

\end{document}